\documentclass[10pt,journal,compsoc]{IEEEtran}
\usepackage[utf8]{inputenc} 
\usepackage[T1]{fontenc}  

\usepackage[breaklinks=true,
            colorlinks,
            linkcolor = red,
            urlcolor  = purple, 
            citecolor = blue,
            bookmarks = black]{hyperref}  
\usepackage{url}  
\usepackage{amssymb}   
\usepackage{booktabs} 
\usepackage{amsfonts}  
\usepackage{nicefrac}  
\usepackage{microtype}  
\usepackage{wrapfig}
\usepackage{bbding}
\usepackage{appendix}
\usepackage{graphicx}
\usepackage{amsmath}
\usepackage{amsthm}
\usepackage{subcaption}
\usepackage{float}
\usepackage{appendix}
\usepackage{xcolor}
\usepackage{multirow}

\usepackage{bm}
\usepackage{bbm}
\usepackage{makecell}
\usepackage{mathtools}
\usepackage{algorithmic}
\usepackage{algorithm}
\usepackage[algo2e,linesnumbered]{algorithm2e}

\ifCLASSOPTIONcompsoc
\usepackage[nocompress]{cite}
\else
\usepackage{cite}
\fi

\ifCLASSINFOpdf
\usepackage{graphicx}
\usepackage{amsmath,amssymb}
\usepackage{color}
\usepackage{multirow}
\usepackage{booktabs}
\usepackage[T1]{fontenc}
\usepackage{float}
\usepackage{color}
\usepackage{setspace}
\usepackage{url}
\usepackage{makecell}
\usepackage{cleveref}

\renewcommand{\paragraph}[1]{\vspace{.25em}\noindent\textbf{#1.}}

\begin{document}

\title{ManiDext: Hand-Object Manipulation Synthesis via Continuous Correspondence Embeddings and Residual-Guided Diffusion}

\author{Jiajun Zhang,
        Yuxiang Zhang,
        Liang An,
        Mengcheng Li,
        Hongwen Zhang,
        Zonghai Hu, 
        \\
        and Yebin Liu,~\IEEEmembership{Member,~IEEE}
\IEEEcompsocitemizethanks{
\IEEEcompsocthanksitem Jiajun Zhang, Zonghai Hu are with the School of Electronic Engineering, Beijing University of Posts and Telecommunications, Beijing 100876, China. E-mail: \{jiajun.zhang, zhhu\}@bupt.edu.cn
\IEEEcompsocthanksitem Yuxiang Zhang, Liang An, Mengcheng Li and Yebin Liu are with the Department of Automation, Tsinghua University, Beijing 100084, China. E-mail:\{yx-z19, li-mc18\}@mails.tsinghua.edu.cn; \{anliang, liuyebin\}@mail.tsinghua.edu.cn 
\IEEEcompsocthanksitem Hongwen Zhang is with the School of Artificial Intelligence, Beijing Normal University, Beijing 100875, China. E-mail: zhanghongwen@bnu.edu.cn
\IEEEcompsocthanksitem Corresponding author: Yebin Liu
}}

\IEEEtitleabstractindextext{%
	\begin{abstract}
        Dynamic and dexterous manipulation of objects presents a complex challenge, requiring the synchronization of hand motions with the trajectories of objects to achieve seamless and physically plausible interactions.
        In this work, we introduce ManiDext, a unified hierarchical diffusion-based framework for generating hand manipulation and grasp poses based on 3D object trajectories. 
        Our key insight is that accurately modeling the contact correspondences between objects and hands during interactions is crucial.
        Therefore, we propose a continuous correspondence embedding representation that specifies detailed hand correspondences at the vertex level between the object and the hand.
        This embedding is optimized directly on the hand mesh in a self-supervised manner, with the distance between embeddings reflecting the geodesic distance.
        Our framework first generates contact maps and correspondence embeddings on the object's surface.
        Based on these fine-grained correspondences, we introduce a novel approach that integrates the iterative refinement process into the diffusion process during the second stage of hand pose generation. 
        At each step of the denoising process, we incorporate the current hand pose residual as a refinement target into the network, guiding the network to correct inaccurate hand poses. 
        Introducing residuals into each denoising step inherently aligns with traditional optimization process, effectively merging generation and refinement into a single unified framework.
        Extensive experiments demonstrate that our approach can generate physically plausible and highly realistic motions for various tasks, including single and bimanual hand grasping as well as manipulating both rigid and articulated objects. 
        Code will be available for research purposes. Project page: \href{https://jiajunzhang16.github.io/manidext}{https://jiajunzhang16.github.io/manidext}.
	\end{abstract}
	\begin{IEEEkeywords}
		 Hand-Object Manipulation, Motion Synthesis, Human-Object Interaction.
\end{IEEEkeywords}}

\maketitle

\IEEEdisplaynontitleabstractindextext

%
\IEEEpeerreviewmaketitle


\IEEEraisesectionheading{\section{Introduction}\label{sec:introduction}}

\begin{figure*}[htbp]
    \includegraphics[width=1.0\textwidth]{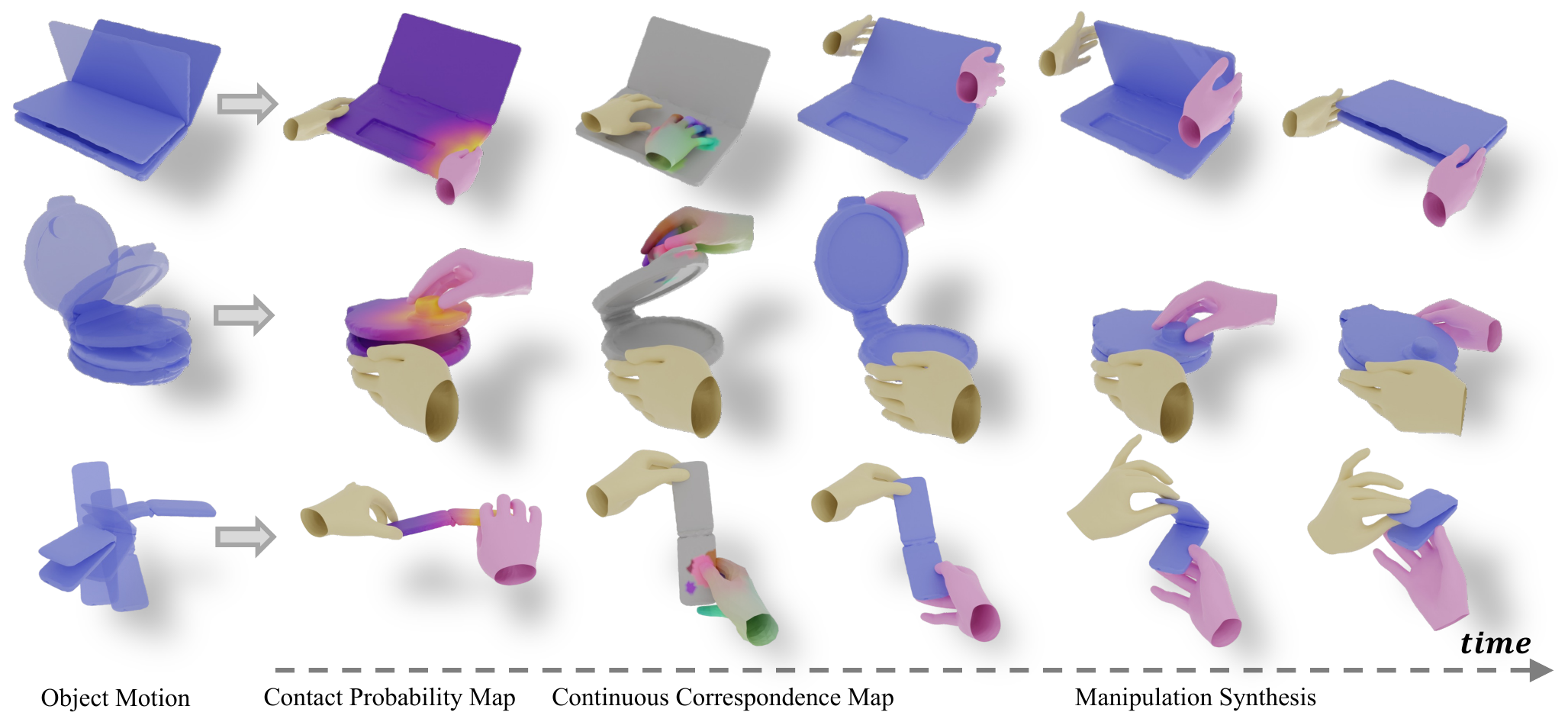}
    \caption{We present ManiDext, which takes a sequence of object motions as input and generates dexterous bimanual hand manipulations. The motions are depicted through snapshots at several key frames. Our hierarchical, diffusion-based pipeline first generates contact probability maps and continuous correspondence maps on the object's surface. These contact details then guide the subsequent stage of hand pose generation.}
    \label{fig:teaser}
\end{figure*}

\IEEEPARstart{T}{he} remarkable dexterity and adaptability of human hands facilitate dynamic and complex manipulations of objects in everyday life.
Dexterous manipulation synthesis of hand poses has significant downstream applications in
virtual reality~\cite{efficientvr,wu2020hand,dexhand}, robotics~\cite{imitationrobot, deeprobotic}, human-computer interaction~\cite{ueda2003handhci}, and embodied AI~\cite{see2touch,bi-dexhands, unidexgrasp++, HInDex}. 
Generating realistic hand pose sequences is a long-standing and challenging problem for two reasons. 
First, the innate dynamics of different objects vary significantly, leading to distinct hand manipulation skills that are difficult to encode in a single neural network. 
Second, it is non-trivial to ensure that the generated hand motions mimic the motor skills of human hands in a physically plausible manner. 
For instance, when opening a laptop with both hands, humans can effortlessly apply appropriate force on the surface, while virtual hands exhibit unnatural motions and may penetrate or be distant from the surface, which violates physical laws.

With the recent advancements in deep learning and the emergence of large-scale motion capture datasets for hand-object interaction~\cite{arctic, hoi4d, obman, fhb, oakink2, ho3d, OakInk, taco, manus, chord, grab}, data-driven generative methods have significantly advanced the field. However, most of these works focus on static hand grasps~\cite{grab, contact2grasp, contactopt, graspTTA,contactgen, graspingfield, halo}, with a few addressing dynamic hand grasp or manipulations~\cite{contactgrasp, manipnet, text2hoi, cams, imos, interhandgen, dgrasp, artigrasp}. These dynamic approaches typically require strong priors or inputs, such as wrist poses, body poses, grasp references, text, or specific object categories, which limit their application scenarios. 

In this paper, we tackle a new and more flexible task of generating dexterous hand manipulations conditioned solely on the object's trajectory. This not only simplifies model inputs but also broadens its application scenarios. This task remains largely unexplored, yet it is crucial for advancing natural and versatile hand-object manipulations. The main difficulties in this task are twofold: first, accurately modeling the interaction regions between the hand and the object during manipulation; second, optimizing to achieve a physically plausible motion sequence with appropriate contact, minimal penetration, and smooth transitions.

To address these challenges, we first propose a continuous correspondence embedding that models hand-object contact state more accurately than contact probability maps or discrete hand part labels used in previous work. Second, we introduce a residual-guided diffusion module that uses hand pose residuals as conditions to guide the generation process. Both of the two modules substantially enhance the performance and efficiency of the generation. Our hierarchical framework first generates contact information on the object’s surface based on its trajectory, and then synthesizes hand poses conditioned on the generated contact information.

Specifically, in the first stage, we reason about contact information through two types of representations: a contact probability map and a continuous correspondence map, as shown in Fig.~\ref{fig:teaser}. 
The contact probability map indicates the likelihood of contact between the hand and the object, while the correspondence embedding provides richer semantic information by detailing the correspondence between object vertices and specific hand vertices.
Specifically, we associate each vertex of the canonical hand model MANO~\cite{mano} with an embedding vector, which is derived through self-supervised optimization. 
The distance between these vectors reflects the geodesic distance between the corresponding vertices on the hand mesh. 
This continuous embedding provides a deformation-invariant point identity over the hand mesh manifold. 
We assigned hand embeddings to each vertex on the object by identifying the nearest vertex on the object's surface to the hand.
Based on these embedding vectors, we use a diffusion-based network to generate both traditional contact probability maps and our continuous correspondence maps on the object's surface, conditioned on the object's trajectories. 
This information accurately indicates detailed contact correspondences between the hand and object, providing stronger guidance for hand pose generation in the subsequent stage.
 
In the second stage, directly using the contact information to generate hand poses may result in artifacts. Previous methods address this by employing explicit optimization or neural refinements subsequently. However, explicit optimization is fragile and may fail to ensure temporal smoothness, while neural refinement requires additional network training.
In our observation, traditional optimization methods iteratively reduce error through gradient descent, which inherently aligns with the step-by-step denoising process in the diffusion models~\cite{ddpm}. Recognizing this natural alignment, we introduce a residual-guided diffusion module, where the hand pose residual errors at the current denoise step serve as additional conditions. 
Specifically, the contact information allow us to establish dense correspondence between object vertices and hand vertices, enabling the computation of geometric residual errors in the current denoising state.
These residual errors, which encode the optimization direction, are included in the diffusion model as conditions. Through the iterative denoising steps of the diffusion process, these errors are continuously updated and gradually minimized, akin to an explicit optimization or neural refinement process. This innovative integration ensures that our second stage benefits from both generation and subsequent optimization in a unified process.

We perform both quantitative and qualitative experiments on various datasets, including ARCTIC~\cite{arctic}, GRAB~\cite{grab} and HOI4D~\cite{hoi4d}, covering scenarios of single-hand and bimanual-hand interacting with both rigid and articulated objects. Comprehensive ablations are presented to support the effectiveness of our proposed continuous correspondence embedding and residual-guided diffusion. The results demonstrate that our method generates high-quality, smooth, dynamic, and physically plausible manipulation poses across a wide range of hand-object interaction patterns. The contribution of this work are summarized as follows: 
\begin{itemize}
\item ManiDext, a diffusion-based framework that is the first to address bimanual hand manipulation synthesis conditioned solely on object trajectories.
\item A novel continuous correspondence embedding representation that accurately models intricate hand-object interaction correspondences.
\item A residual error guided module in the diffusion denoising step, seamlessly integrating generation and refinement processes.
\end{itemize}
\def\etal{\emph{et al}.}
\section{Related Work}\label{sec:related_work}

\subsection{Static Hand Grasp Synthesis} 
Generating realistic hand grasps remains challenging due to the complex geometry of objects and the intricate constraints of hand articulation and contact dynamics. Over the past decades, there have been numerous attempts to use physical control methods~\cite{pollard2005physically, kry2006interaction, li2007data} for hand-object manipulation synthesis. These methods are sensitive to initialization and struggle to generalize across different objects with fixed hyperparameters. 
To enable data-driven synthesis instead of physical control, Taheri \etal~\cite{grab} first collect a comprehensive dataset of hand grasps, and generate plausible grasping pose via a coarse-to-fine process. 
Subsequent research mostly focuses on how to improve the representation of contact and effectively utilize contact information.
Karunratanakul \etal~\cite{graspingfield} proposed an implicit representation for modeling hand object relation based on signed distance fields. 
Grady \etal~\cite{contactopt} improves hand pose estimation by optimizing hand-object contact states.
Jiang \etal~\cite{graspTTA} improves grab synthesis by modeling the consistency between the hand contact points and object contact regions.
While contact maps are effective in modeling the contact regions, they fall short in providing semantic information, particularly in identifying which specific parts of the hand are involved in the contact.
Liu \etal~\cite{contactgen} utilizes three types of maps: contact map, hand part map and direction map to describe contact information, guiding the synthesis of hand grasps.  
However, hand part labels still have limitations in continuity and semantic richness, failing to capture finer-grained contact information, accurately reflect specific parts of hand-object contact, and comprehensively describe complex hand-object interactions.
In addition to utilizing contact information, 
Karunratanakul \etal~\cite{halo} introduces a neural implicit surface representation for human hands driven by keypoint-based skeleton articulation.
Lee \etal~\cite{interhandgen} learns a prior of bimanual by dropout one hand and mask another. 
From previous work, it is widely acknowledged that reasoning contact information is crucial, as it forms the foundation for accurate hand-object interactions.
In this work, the proposed continuous embedding provides dense correspondence, mapping vertices on the object to vertices on the hand, and is superior to the commonly used representations of contact maps and hand part labels. This embedding is discriminative, offering better distinction for nearby points by reflecting geodesic distance rather than Euclidean distance, which enhances the pose generation process. Additionally, this representation can be seamlessly integrated into previous works.

\subsection{Dynamic Hand Manipulation Synthesis}
Dynamic dexterous manipulation of objects is a more challenging task that necessitates synchronizing the generated motions of the hand with the motion trajectory of the object to achieve seamless and physically plausible manipulation. 
Previous work has explored various approaches to address this challenge. 
Zhang \etal~\cite{manipnet} utilize multiple sensors to model the spatial relationship between hand and objects. 
Zheng \etal~\cite{cams} introduced a novel canonicalized contact space modeling approach along with an optimization-based refinement process, enhancing the accuracy and realism of hand-object interactions. 
Taheri \etal~\cite{grip} achieves more stable grasps through temporal consistency and customized sensor modeling of hand-object spatial relations.
However, the setting of these works are relatively simplified compared to ours, which solely conditions on the object's trajectory. Some condition on wrist and object trajectories~\cite{manipnet}, others on body pose and object trajectories~\cite{grip}, and some decompose the motion sequence into multiple steps~\cite{cams} with relatively simple actions and trained at the category level. 
Furthermore, most of these works primarily focus on grasping, with little to no emphasis on, or only very simplistic, manipulations. 
Additionally, these methods often require an extra post-processing module, either to establish correspondence for explicit optimization or to apply neural refinement, which is time-consuming and fragile. 
There are also studies~\cite{toch, geneoh} specifically focused on refining incorrect hand-object interaction sequences to ensure better temporal consistency and interaction generation. 
Besides deep learning based methods, methods using reinforcement learning (RL) combined with physical simulation to learn effective manipulations policies have been proposed~\cite{dgrasp, artigrasp}. However, these learned policies are often constrained to the specific types of objects and manipulations.
From previous work, it is evident that establishing the relationship between objects and hands over time, as well as incorporating refinement modules in post-processing, enhances the quality of motion synthesis. 
From this perspective, we introduce the residual-guided diffusion module, which establishes hand-object residuals over time sequences and use these as conditions to effectively integrates the generation and refinement modules from previous work into a single unified module.

\begin{figure*}[!t]
    \centering
    \includegraphics[width=1.0\linewidth]{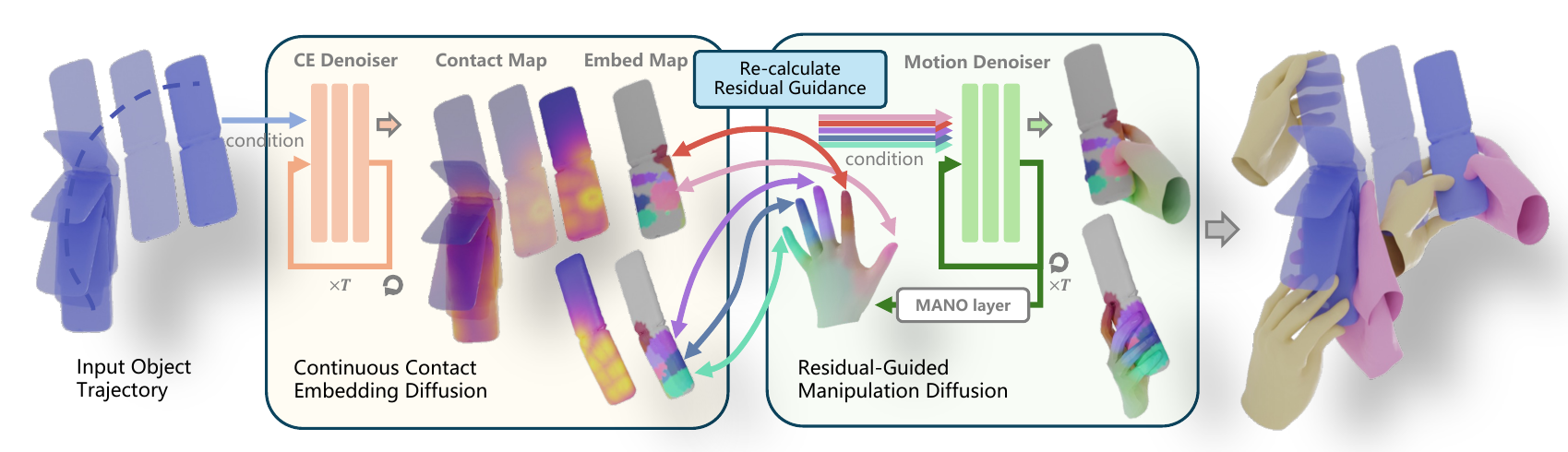}
    \caption{Method Overview. Given a sequence of object motion trajectory, we adopt a hierarchical diffusion-based framework to gradually generate the hand poses that manipulate the object. First, we generate contact information on the object's surface, which includes a contact probability map and a continuous correspondence embedding map. These information provides dense correspondence, allow us to compute the geometric residual error at each diffusion timestep $T$. Subsequently, we use the generated contact information and the computed residual error as conditions to generate the manipulation hand poses.}
    \label{fig:pipeline}
\end{figure*}

\subsection{Diffusion Model in HOI Synthesis}
Diffusion models have recently gained attention for their ability to generate high-fidelity data in areas such as images~\cite{ddpm}, videos~\cite{videodiffusionmodel}, motion sequences~\cite{mdm, mld}, text~\cite{textdiffusion}, and audio~\cite{audiodiffusion}.
This versatility has inspired research into using diffusion models for HOI synthesis. 
For instance, Xu \etal~\cite{interdiff} predicted future object and body motions based on past observations. Li \etal~\cite{omomo} generated body motions conditioned on object motions.
There are also many works that generate human~\cite{cghoi, chois, hoidiff} or hand-object~\cite{text2hoi} interactions based on textual descriptions.
These examples highlight the tremendous potential of diffusion models in generating dexterous hand motions.
To our knowledge, our work is the first diffusion-based method for dexterous manipulation synthesis conditioned on object motion, a field that remains largely unexplored.

\section{Overview}\label{sec:Overview}
An overview of our ManiDext is shown in Fig.~\ref{fig:pipeline}.
Given an object's geometry $M_{\text{obj}}$ and motion trajectory $\psi \in \mathbb{R}^{L \times K}$, our goal is to generate bimanual hand poses $H \in \mathbb{R}^{L \times D}$, where $L$ represents the length of the temporal sequence, and $K$ and $D$ represent the dimensions of the representations for the object and hand motion.
Preliminarily, we introduce the temporal modeling of the object shape, hand poses, and hand-object spatial relationships (Sec.~\ref{sec:Representation}).
Based on these representations, we introduce our hierarchical framework that generates dynamic hand manipulations and decomposes the generation process into two stages. 
In the first stage, we initially introduce how the continuous surface embeddings on the hand are obtained through self-supervised optimization (Fig.~\ref{fig:embedding}). Then, we generate contact map $C_{\text{map}} \in \mathbb{R}^{L \times n \times 2}$ and correspondence embedding map ${E_{\text{map}} \in \mathbb{R}^{L \times n \times 2 \times d}}$ on the object's surface based on the object trajectory (Sec.~\ref{sec:continuous correspondence embedding}).
Here, $n$ denotes the number of sampling points used for encoding the object, $2$ denotes bimanual hands, and $d$ represents the dimension of the continuous surface embeddings. 
In the second stage, based on the generated contact information $I = \{C_{\text{map}}, E_{\text{map}}\}$ and the noised hand poses $\hat{H}_t$, we establish a dense correpondence between the object's surface and hand regions (Fig.~\ref{fig:correspondence}). We then compute the residual error $R_{t} \in \mathbb{R}^{L \times 778 \times 2 \times 3}$ (Fig.~\ref{fig:residual}) at the current denoising step $t$ and use this residual to guide the network's predictions (Sec.~\ref{sec:residual guided diffusion}).
Here, $778$ represents the number of MANO hand vertices, and $3$ represents the delta vectors between each vertex and the corresponding contact point on the object. Finally, we generate the hand pose $H_0$ based on the combined condition $\{\psi, I, R_{t}\}$. 
\section{Representation}\label{sec:Representation}
In this section, we introduce the representations of object shape, hand pose, and hand-object motion temporally. These components form the foundation and serve as inputs for the network to generate dynamic bimanual hand manipulations.
\subsection{Object Shape Representation}
In contrast to methods that require separate training for each object or category, our approach is category-free, meaning that a single generative model is shared across all objects, allowing for greater flexibility and generalization. 
However, different objects have varying shapes and topologies, thus we need a unified topological representation for all objects. Similar to~\cite{omomo, grab}, we adopt basis point set (BPS)~\cite{bps} representation, whose ordered, fixed-length vector encoding is well-suited to our requirements. Specifically, we sample $n$ basis points within a sphere of fixed radius and form the basis point set as $B \in \mathbb{R}^{n \times 3}$, BPS encodes the directional vectors to the nearest points on the object surfaces as
\begin{equation}
\label{eq:bps}
  X^{B} = \{\text{argmin}_{x \in X} d(b_q, x) - b_q\} \in \mathbb{R}^{n \times 3},
\end{equation}
where $X$ is the set of object vertices, $n$ is the number of basis points, $b_q$ represents the $q_{th}$ basis points and $X^{B}$ represents the features after BPS encoding. 
\subsection{Hand Pose Representation}
Hand poses are represented as $H = \{H_{\text{left}}^{l}, H_{\text{right}}^{l}\}_{l=1}^{L}$, where $L$ denotes the sequence length. For the bimanual hand, $H_{\text{left}}^{l} \in \mathbb{R}^{99}$ and $H_{\text{right}}^{l} \in \mathbb{R}^{99}$ are both composed of a 99-dimensional vector by flattening and concatenating the 3D hand translation $\text{Trans}^{l} \in \mathbb{R}^{3}$, root orientation $r_{root}^{l} \in \mathbb{R}^{1 \times 6}$ and hand poses $\theta^{l} \in \mathbb{R}^{15 \times 6}$ represented in rotation 6D~\cite{pose6d}. We adopt parametric model MANO~\cite{mano} to reconstruct bimanual hand meshes.

\begin{figure}
    \centering
    \includegraphics[width=1.0\linewidth]{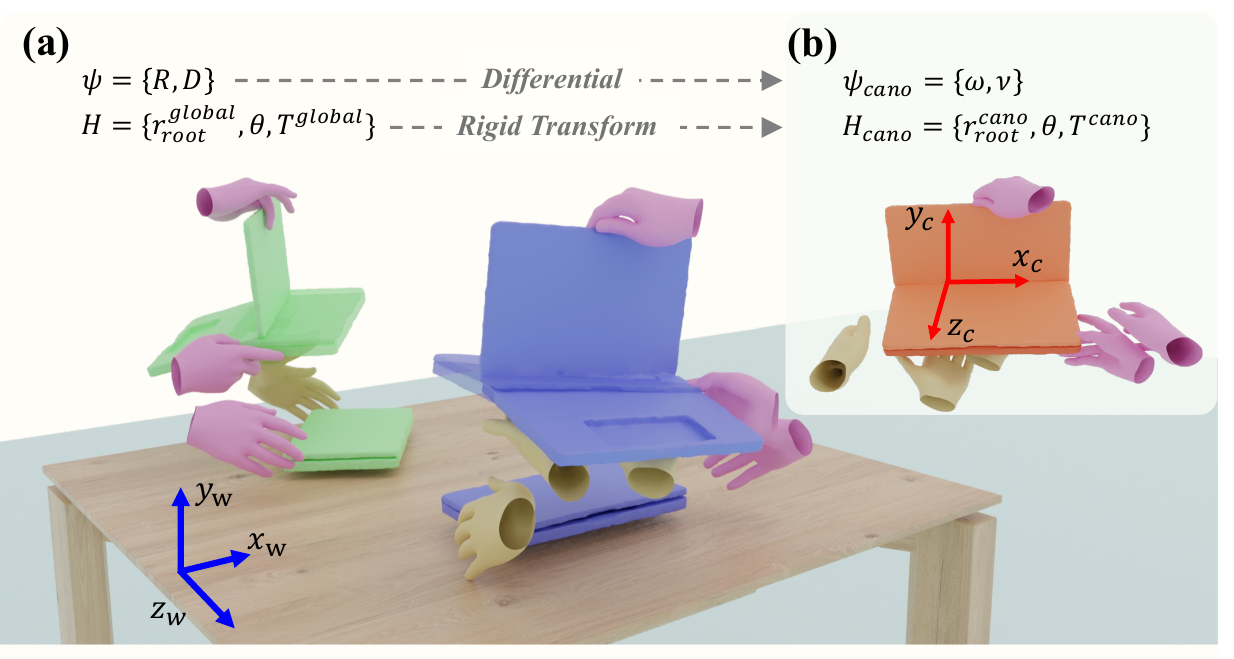}
    \caption{Illustration of differential object-centric motion modeling.
    (a) shows the motion modeling in the world coordinate system. The \textbf{blue} shows the object's position and orientation during data collection, while the \textbf{green} represents the same relative motion from a different orientation.
    (b) shows the hand-object modeling in object's canonical coordinate system, eliminating global information, shown in \textbf{orange}.}
    \label{fig:world2cano}
\end{figure}

\subsection{Differential Object-Centric Motion Modeling}
\label{sec:object-centric-model}
In hand-object interaction modeling, object dynamics are typically represented via their CAD models in a canonical space, capturing global rotations and translations, while hands are modeled using parametric models. However, this global configuration often leads to overfitting to training poses and poor generalization to unseen object trajectories.
For example, in the ARCTIC dataset~\cite{arctic}, data collection is performed on a table with all participants interacting with objects from the same direction (Fig.~\ref{fig:world2cano}(a), Blue). When testing with interactions such as opening a laptop from the back or side of the table (Fig.~\ref{fig:world2cano}(a), Green), the network struggles to generalize to unseen trajectory. Our observation is that despite changes in the global position and orientation of the laptop across these scenarios, the relative motion between the hand and the laptop, such as the angle and direction of opening, remains consistent. 
By focusing on these consistent relative motions instead of the global properties, the network better learns the underlying mechanics of hand-object interaction.
Since the global trajectory is given in our task, we introduce a differential object-centric modeling approach (Fig.~\ref{fig:world2cano}(b), Orange) to describe the hand-object interaction within the object's canonical coordinate system, effectively eliminating the influence of global information.

Specifically, given a sequence spanning from $l = 0$ to $L$, the object's global motion is represented by $\psi_{\text{world}}^{l} = \{R_l, D_l\}$ and the hand's pose by $H_{\text{world}}^l = \{r_{\text{root}}^{\text{global}}, \theta, T^{\text{global}}\}$. To eliminate the influence of global information, we redefine the motion in the object's canonical coordinate system as $\psi_{\text{cano}}^{l} = \{\boldsymbol{\omega}_l, \mathbf{v}_l\}$ and $H_{\text{cano}}^l = \{r_{root}^{\text{cano}}, \theta, T^{\text{cano}}\}$.
Here, $\boldsymbol{\omega}_l$ denotes the object's angular velocity, computed as $\boldsymbol{\omega}_l = {R_{l-1}}^{T}\cdot R_l$, and $\mathbf{v}_l$ represents the object's translation velocity, defined as $v_l = {R_{l-1}}^T\cdot(D_l - D_{l-1})$. 
Notably, for articulated objects, an articulation angular velocity $\boldsymbol{\alpha}_l$ is also included.
These factors capture the local rotation and translation relative to the previous frame, fully characterizing the object's motion dynamics. 
Correspondingly, the hand’s motion is also transformed into the object’s canonical coordinate system, as $r_{\text{root}}^{\text{cano}} = R^{T} \cdot r_{\text{root}}^{\text{global}}$ and  $T^{\text{cano}} = R^{T} \cdot (T^{\text{global}} - D)$.
Our training is conducted in the object's canonical space, using $\psi_{\text{cano}}$ as the input condition to predict $H_{\text{cano}}$. This velocity-based description of the object's dynamics serves as a differential representation, allowing us to recover the object's rotation and translation at each frame by integrating the angular and translation velocities.  
When the object's initial pose is given, we can easily transform the canonical representation back to the world coordinate system.

\section{Method}
Our hierarchical generation framework is entirely based on diffusion. In this section, we first introduce the foundational concepts of diffusion as a preliminary. We then detail our contributions at each stage: the continuous correspondence embedding and the residual-guided diffusion module. 
\subsection{Conditional Denoising Diffusion}
Our two-stage generation process, comprising contact information generation and hand pose generation, is entirely based on the diffusion model. 
The denoising diffusion model~\cite{ddpm} is a probabilistic generative model that consists of a forward process and a reverse process. The forward process converts the original data representation $x_{0}$ by gradually adding Gaussian noises for $T$ step with predefined means and variances $\beta_{t}$ in a Markov Chain, 
\begin{equation}
\label{eq:forward_diffusion_step}
    p(x_{t}|x_{t-1}) := \mathcal{N}(\sqrt{1-\beta_{t}}x_{t-1}, \beta_{t}I),
\end{equation}
The reverse process is to generate desired data $p(x_0)$ from random noised sample $x_{T} \sim \mathcal{N}(0,I)$, defined as 
\begin{equation}
\label{eq:reverse_diffusion_step}
    p_{\theta}(x_{t-1}|x_{t}, c) := \mathcal{N}(\mu^{\theta}_{t}(x_t, t, c), \sigma_{t}^{2}I),
\end{equation}
where $c$ represents the conditions for generation and $\sigma_{n}$ is determined variances. While $\mu_t$ is untrackable, it is predicted from a neural network $\mu_t^{\theta}(x_t, t, c)$. 
In the reverse process, we follow~\cite{mdm} to predict the clean motion $x_{0}$ instead of noise in order to incorporate geometry constraints. 

\subsection{Continuous Correspondence Embedding}
\label{sec:continuous correspondence embedding}
Previous approaches~\cite{contactopt, graspTTA, grab, text2hoi} use a contact map applied to object point clouds to guide the hand pose generation, where each point $ p \in X^{B} $ in the point set  $X^{B}$ is associated with a contact probability ranging from 0 to 1, denoted as $p_{contact} \in [0,1]$. Additional efforts, such as those in~\cite{contactgen, graspingfield}, utilize labels for corresponding parts of the hand, denoted as $p_{class} \in  [0, \ldots, \chi] $, where $\chi$ represents predefined parts of the hand. Despite these efforts, they still struggle to provide an accurate contact correspondence between points on the object and vertices of the hand. 

\begin{figure}
    \centering
    \includegraphics[width=1.0\linewidth]{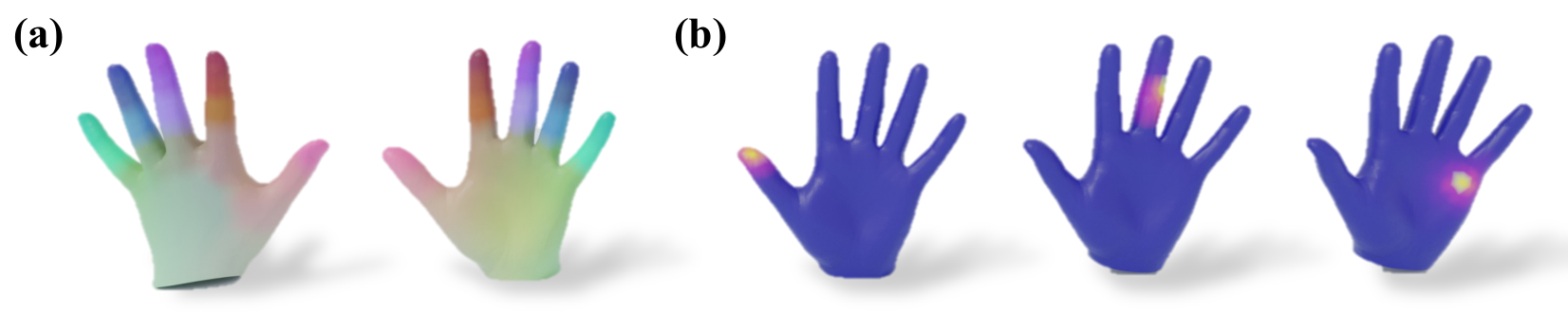}
    \caption {(a) Visualization of the optimized continuous surface embedding of the hand in three dimensions, represented with colors. (b) Visualization of the embedding distances between a query vertex and all other vertices.}
    \label{fig:embedding}
\end{figure}

Inspired by Continuous Surface Embedding~\cite{cse}, we propose to employ a specialized embedding to model a more dense, precise and detailed correspondence between object points and hand vertices. 
In contrast to Continuous Surface Embeddings (CSE), which are optimized concurrently with the network during training, our proposed continuous correspondence embeddings are pre-optimized offline through a self-supervised approach.
Let $S \in \mathbb{R}^{778 \times 3}$ be the set of vertices from MANO hand model in the canonical pose, for each vertex $ s_i \in S $, we compute the geodesic distances to all other vertices, yielding a distance matrix $G \in \mathbb{R}^{778 \times 778}$, where each element $G_{ij}$ represents the geodesic distance between vertex $s_i$ and vertex $s_j$. 
The continuous correspondence embedding is represented as $ E \in \mathbb{R}^{778 \times d}$, where $d$ denotes the dimension of embedding. 
We optimize $E$ by minimizing the $\mathcal{L}_{bce}$ distance between $\Phi_{emd}$ and the normalized geodesic distance $\Phi_{gt}$ as follows:
\begin{equation}
\label{eq:embedding}
    \Phi_{emd} = \exp\left(-\|\mathbf{E}_i - \mathbf{E}_j\|\right), 
    \vspace{10pt}
    \Phi_{gt} =  \exp\left(-\frac{G^2}{2\sigma^2}\right),
\end{equation}
\begin{equation}
\label{eq:embeddingloss}
    \mathcal{L}_{embedding} = \mathcal{L}_{bce}\left(\Phi_{emd}, \Phi_{gt}\right).
\end{equation}
As shown in the Fig.~\ref{fig:embedding}(a), we visualize the embeddings on MANO surfaces, where the embeddings are in three dimensions and normalized for visualization.
The results appear smooth and continuous, showing similarities between adjacent regions while being highly discriminative between distant regions. 
To verify the discrimination of the trained embedding, in Fig.~\ref{fig:embedding}(b), given query points on the thumb, middle finger, and palm, we visualize $\Phi_{emb}$ of each query vertex, which represents the similarity between the query point and other points.
Due to the identical topological structure of MANO for both the left and right hands, we apply the same embedding to both hands. 

\begin{figure}
    \centering
    \includegraphics[width=1.0\linewidth]{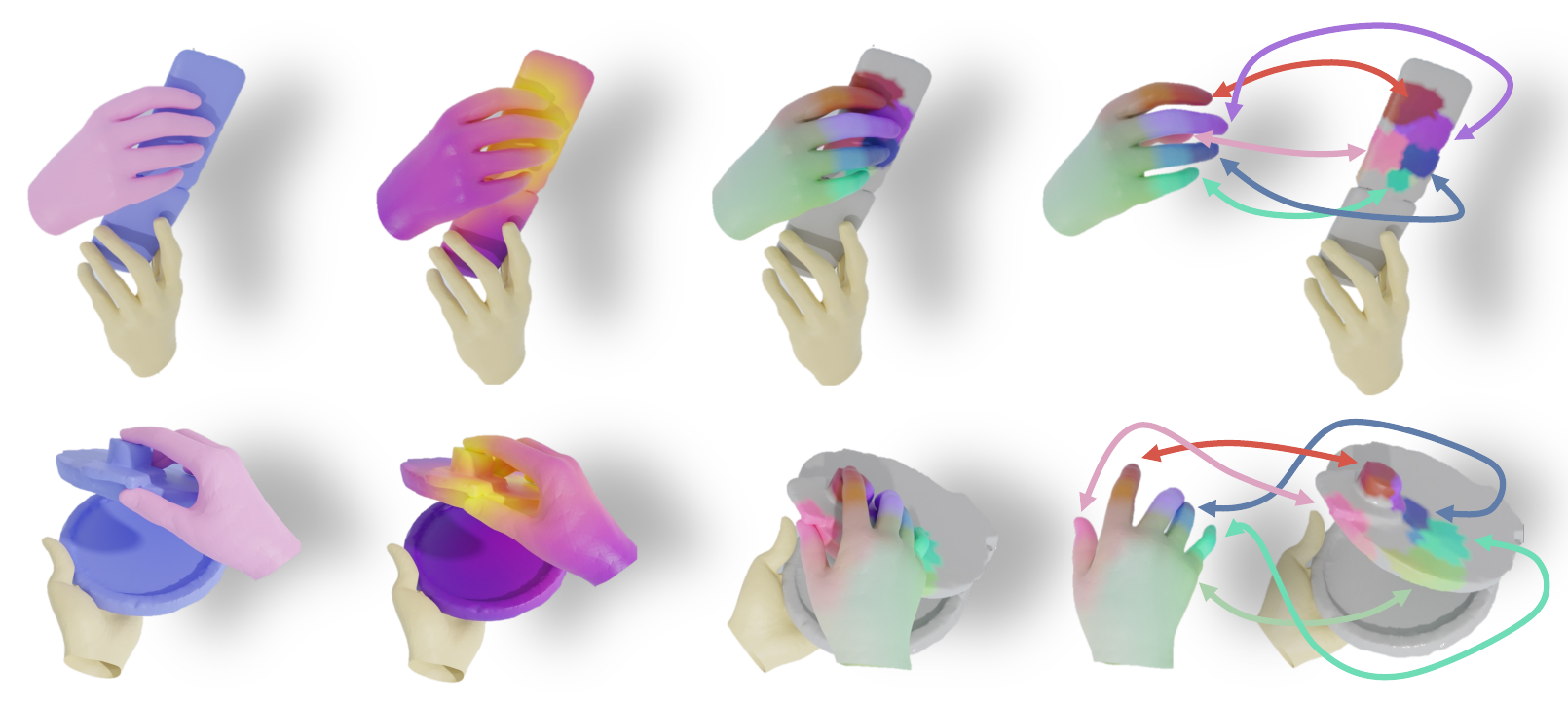}
    \caption{Illustration of contact probability and correspondence embeddings defined on both the object and hand surfaces.}
    \label{fig:correspondence}
\end{figure}

In the first stage, given an object and its trajectory, we use a diffusion framework to generate an object-centric contact representation $I$. Our object contact representation $I = \{C_{\text{map}}, E_{\text{map}}\}$ consists of two maps. The contact map  $\bm{C_{\text{map}}} \in \mathbb{R}^{L \times n \times 2}$ indicates the contact probabilities of each sampled vertex on the object with the left and right hand, respectively.
The correspondence embedding map $\bm{E_{\text{map}}} \in \mathbb{R}^{L \times n \times 2 \times d}$ provides detailed semantic information by indicating which specific regions of the hand are in contact with each sampled vertex on the object's surface.
During data preprocessing, for each vertex on the object, we find the nearest hand vertex $i$ and assign its embedding $E_i$ to construct the $E_{\text{map}}$. The distance between them is then normalized into contact probability to construct the $C_{\text{map}}$, as shown in Fig.~\ref{fig:correspondence}.
The process of contact information generation follows the diffusion process~\cite{ddpm}. Our training target is denoted as $z_0 = I \in \mathbb{R}^{L \times n \times 2 \times (1 + d)}$, where $L$ represents the temporal length, $n$ represents the predefined number of BPS points, 2 denotes the bimanual hand and the last dimensions includes a single contact probability and $d$ dimensions for embeddings. We adopt transformer~\cite{transformer} architecture as denoising network $\mathcal{G}$, formally,  
\begin{equation}
    \mathbf{\tilde{z}} = \mathcal{G}(\mathbf{z}_t, t, \psi),
\end{equation}
where $\psi = \{X^{B}, \boldsymbol{\omega}, \mathbf{v}\}$ is the conditions of object motions for the entire sequence, as described in Sec.~\ref{sec:object-centric-model}. The training process is supervised by the objective function as: 
\begin{equation}
    \mathcal{L} = \lambda_{\text{contact}} \mathcal{L}_{bce}(C_{\text{map}},  \hat{C_{\text{map}}}) + \lambda_{emb}|| E_{\text{map}} - \hat{E_{\text{map}}} ||_2,
\end{equation}
where $\hat{\text{hat}}$ represents the ground truth, and $\lambda$ denotes the coefficient of each loss function. Notably, we supervise the contact map across all vertices, but for the embeddings, supervision is applied only to vertices where the ground truth contact probability is greater than 0.5. 

\subsection{Residual Guided Diffusion}
\label{sec:residual guided diffusion}
So far, we have generated a contact map and a correspondence map, providing essential spatial and contextual information for guiding hand pose generation. 
However, directly using these conditions may lead to artifacts like penetration, not in contact and motion inconsistency. 
To address these, previous work introduce a post-processing step, using either neural refinement~\cite{text2hoi, grip} or optimization-based method~\cite{contactopt, cams, manipnet}.
These methods are often time-consuming and unstable, requiring additional networks or explicit optimization. 
Our insight is that these iterative refinement approach align well with the gradual denoising concept of diffusion processes. 
The reverse diffusion process gradually refines random noise hand poses $z_t$ into clean samples $z_0$, serving as a form of iterative refinement.
Given the contact and correspondence maps generated in the first stage, we can establish the correspondence between object vertices and hand vertices in contact, and calculate the geometric residuals between them at each reverse diffusion step $t$. The residual information helps the network aware of the current state's error and provides clear guidance for denoising. 
Therefore, we introduce the continuously updated residual information as a condition to guide the denoising process, named Residual Guided Diffusion.
Throughout the reverse process from $T$ to $0$, the residual condition is continually updated and reduced, guiding the direction of the denoising process at each step. 

\begin{figure}
    \centering
    \includegraphics[width=1.0\linewidth]{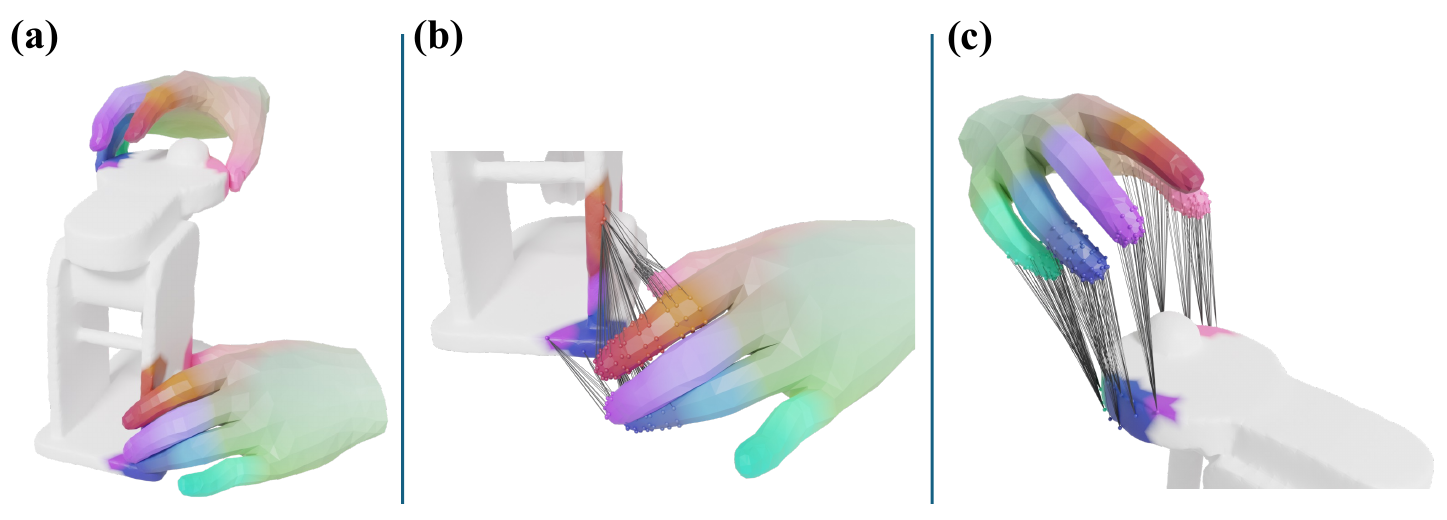}
    \caption{Illustration of the pose residual calculation process. (a) shows the ground truth of hand poses and contact information $I = \{C_{\text{map}}, E_{\text{map}}\}$. (b) and (c) depict the computation of the geometric residual $r_{ij^*}''$ (shown by black lines) for the left and right hands, respectively, in the noised hand pose state $\mathbf{z}_t$.
    }
    \label{fig:residual}
\end{figure}

Specifically, at each reverse diffusion step $t \in \{T, T-1, \ldots, 0\}$, for each hand vertex $h_i$,
we compute its embedding similarity with vertices on the object surface to identify the corresponding object vertex it is most likely to contact, defined as $j^* = \arg\min_j \|E_i - E_{\text{map}}^{j}\| $, where $j^{*}$ is the index of the closest matching object vertex. 
Once $j^*$ is identified, the residual error is calculated as: $r_{ij^*} = h_i - b_{j^*}$, where \(b_{j^*}\) represents the $j$-th encoded object vertex. By iterating over all hand vertices, the residual information $r \in \mathbb{R}^{778 \times 2 \times 3}$ is constructed.   
However, this way of calculation determines a residual for every hand vertex, even though most of these vertices are not in contact with object. Therefore, to focus solely on the contact regions, we apply the contact probabilities as a mask to the residual calculation as: $r_{ij^*}' = C_{map}^{j^{*}} \times (h_i - b_{j^*})$. 
Moreover, due to the downsampling introduced in BPS encoding, multiple hand vertices may correspond to a single object vertex, resulting in a many-to-one mapping, which introduces ambiguity in the correspondence. 
We further address this ambiguity by apply embedding similarity to encode residuals with varying degrees for each hand-object vertex pairs as: $r_{ij^*}'' = \Phi_{emd}^{ij^{*}} \times C_{map}^{j^{*}} \times (h_i - b_{j^*})$, where $\Phi_{emd}^{ij^{*}} = \exp\left(-\|E_i - C_{map}^{j^{*}}\|\right)$ represents the embedding similarity between the hand vertex $h_i$ and the object vertex $b_{j^*}$. 
Fig.~\ref{fig:residual} provides an intuitive explanation of the residual calculation.
From a higher perspective, the combination of contact probabilities and embedding similarity can be interpreted as forces exerted at each hand vertex. The geometric residual errors, on the other hand, represent the magnitude of displacement needed to correct the hand pose to align accurately with the object. 

In the second stage, our training target is denoted as $z_0 \in \mathbb{R}^{L \times 2 \times 99}$, 
where $L$ represent the sequence length, 2 represents the bimanual hand and 99 represents the concatenation of hand parameters. We adopt the same transformer architecture as in the first stage for denoising network $\mathcal{G}$, formally, 
\begin{equation}
    \mathbf{\tilde{z}} = \mathcal{G}(\mathbf{z}_t, t, (\psi, I, r_{t}'')),
\end{equation}
where $\psi$ is the object motions, $I$ is the contact information generated in first stage and ${r_t}''$ is the residual error condition varied at each diffusion iteration step $t$. The overall objective function for training the second stage is written as: 
\begin{equation}
\label{eq:s2loss}
\begin{split}
    \mathcal{L_\text{total}} & = \lambda_{data} \cdot || z - \hat{z} ||  + \lambda_{pen} \cdot \mathcal{L_\text{pen}}  \\ &  + 
    \lambda_{joints} \cdot || (F(z_{l}) - F(\hat{z_{l}}) || \\ & +
    \lambda_{vel} \cdot || (F(z_{l+1})-F(z_{l})) - (F(\hat{z_{l+1}}) - F(\hat{z_{l}}) )||  \\ & +
    \lambda_{att} \cdot || \Phi_{emb}^{ij^{*}} \cdot C_{map}^{j^{*}} \cdot (h_{i} - \hat{h_{i}}) ||  \\ & +
    \lambda_{consist} \cdot  || C_{map}^{j} \cdot (||h_{i^*} - b_j||_2 - P2D(C_{map}^{j})) ||.
\end{split}
\end{equation}
Here, $\lambda$ represents the loss coefficient, $\hat{\text{hat}}$ denotes the ground truth. $\mathcal{L_\text{data}}$ supervises the simple objective of hand pose and translation parameters. $\mathcal{L_\text{pen}}$ follows~\cite{graspTTA} to prevent the penetration between hand and object meshes. 
$\mathcal{L_\text{joints}}$ and $\mathcal{L_\text{vel}}$ constrains the velocity of joints to ensure temporal stability across sequence and $F$ denotes the forward kinematic function converting MANO parameters to hand joints.
$\mathcal{L_\text{att}}$ denotes the attraction loss, which supervises the alignment between the hand vertices involved in the residual calculation and their corresponding ground truth vertices.
$\mathcal{L_\text{consist}}$ denotes the consistency loss, which ensures that the predicted hand positions are consistent with the contact and correspondence map. For each object vertex $b_j$, we find the closest hand points $i^*$ based on the embedding similarity. The distance between them is then computed to match the ground truth distance, which is derived from the contact probability map using the $P2D$ conversion, defined as $ d= (-2 \cdot \sigma^2 \cdot \log(C_{map}))^{1/2}$. 
\newpage
\section{Experiments}\label{sec:experiments}

\subsection{Dataset}
We evaluate ManiDext on the GRAB~\cite{grab} and ARCTIC~\cite{arctic} datasets. 
The GRAB dataset contains hand poses of 10 subjects grabbing 51 different rigid objects. 
The ARCTIC dataset contains 10 subjects interacting with 11 objects. ARCTIC is more challenging because it involves articulated objects and more complex motions, including grabbing and manipulation. Moreover, GRAB primarily focuses on single-hand grasping, while ARCTIC captures bimanual cooperative actions. For each dataset, we use the object motions from the last subject as the test set and the remaining subjects as the training set, following~\cite{text2hoi}. The training is conducted separately for each dataset from scratch, with all objects being trained in a single model.

\begin{figure*}
     \centering
    \includegraphics[width=1.0\linewidth]{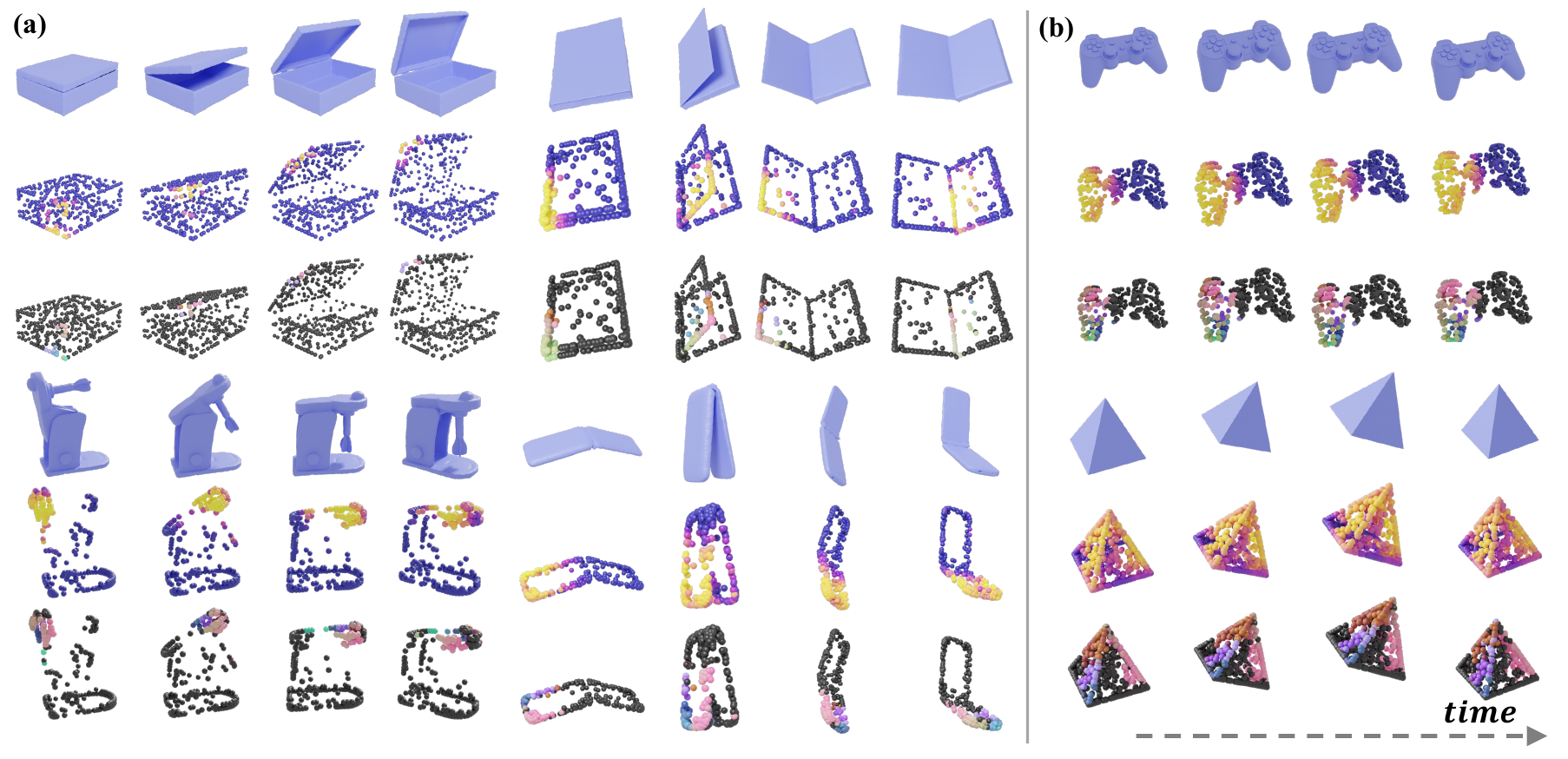}
    \caption{Qualitative results of first stage. We visualize object motion (top row), generated contact map (middle row), and embeddding map (bottom row) from ARCTIC~\cite{arctic} (a) and GRAB~\cite{grab} (b) test sets in each block. 
    The contact map is visualized using the plasma colormap, where warmer colors indicate highter contact probabilities.
    The colors of embedding map correspond to hand embedding shown in Fig.~\ref{fig:embedding}. Our method generates separate contact and embedding maps for both left and right hands, yet we only show those for left hand here for clarity.} 
    \label{fig:s1_results}
\end{figure*}

\subsection{Implementation Details}
We randomly choose 512 points as fixed BPS~\cite{bps} encoding basis points, with a radius of 0.3 meters for ARCTIC and 0.1 meters for GRAB. For each frame, we compute the minimum bounding sphere of the object and use its center as the center for the BPS points. For both datasets, we clip out the silent postures at the beginning and end of the raw sequences based on contact information, ensuring that only valid data is used for training. For each sequence during training, we randomly sample a start frame and set the sequence length to 120 frames, padding with zeros for shorter sequences. To mitigate the impact of errors from the first stage and increase robustness, we mask all contact information with a 20\% probability when training the second stage. For the remaining 80\%, we randomly mask the contact information for 30\% of the frames within each training sequence. 

Both stages of the denoising network share the same transformer architecture, featuring 4 self-attention blocks and 4 attention heads, differing only in the input condition encoders, which are implemented using distinct Multi-Layer Perceptrons (MLP).
For the diffusion module, we follow the settings of MDM~\cite{mdm}, using $T=1000$ noising steps and predicting the signal $z_0$ directly, which facilitates the computation of geometric loss.
We use Adam optimizer, with a constant learning rate of $1 \times 10^{-4}$ and a batch size of 32. Each stage is trained for 300k steps on a single RTX 3090 GPU, taking approximately 3 days per stage. To accelerate training, we only introduce penetration loss $\mathcal{L_\text{pen}}$ after the 250k step.

\subsection{Evaluation Metrics}

\paragraph{Inter-penetration Volume (Pen)} 
Ideally, there should be no physical overlap between hands and objects. To assess the physical plausibility of our generated results, we compute the interpenetration volume (cm$^3$) at each time step by voxelizing the hand and object meshes into 5mm$^3$ cubes and measuring the overlapping voxels, following ~\cite{graspTTA, contactgen}.

\paragraph{Valid-Contact Ratio (V-Contact Ratio)}
Following the methods of ~\cite{contactopt,graspTTA, contactgen, halo}, we evaluate the contact ratio to calculate the proportion of manipulations that are in contact with objects. Given that some bimanual manipulations in ARCTIC can be performed by a single hand, potentially leaving the other hand far from contact, we specifically measure the contact ratio only for cases where the maximum contact probability generated in the first stage is greater than 0.8, effectively measuring precision, as it compares the predicted contacts with the actual contacts.

\paragraph{Valid-MPVPE (V-MPVPE)} 
To validate how well our generated hand poses align with the contact information used to guide the generation process, we use the ground truth contact information to evaluate the second stage and measure the Mean Per-Vertex Position Error (MPVPE) in centimeters (cm) for hand regions where the contact probability in the ground truth is greater than 0.8.

\paragraph{Perceptual Score (P-Score)} 
We conducted a user study following~\cite{cams, contactgen, graspTTA} to assess the perceptual quality of the generated hand poses. Specifically, we invited 10 participants who were not familiar with this area of research. Each participant evaluated the generated hand poses across three dimensions: smoothness, physical plausibility, and alignment with the object's motion. Ratings were given on a scale of 1 to 5, where 1 represents the lowest quality and 5 represents the highest. For each object in the test set, we randomly selected two motion sequences, and report the mean rating for each method.

\subsection{Contact and Correspondence Map Generation}
Many related hand synthesis works, such as ContactGen~\cite{contactgen}, ContactOpt~\cite{contactopt}, Text2HOI~\cite{text2hoi}, etc., utilize a two-stage framework that first generates contact information and then generates the hand poses. 
However, quantitative evaluation of the generated contact information is inherently challenging, as generative model can produce a wide range of plausible contact information, making it difficult to assess them using objective metrics. 
This is also an issue that previous works have struggled with.
Therefore, we focus on qualitative evaluation, providing visualizations to demonstrate the quality and plausibility of the generated contact maps. 
As shown in Fig.~\ref{fig:s1_results}, we present the object's motion (first row), the contact maps generated on the sampled points after BPS encoding (second row), and the correspondence maps (third row). 
For the manipulation task from the ARCTIC dataset in Fig.~\ref{fig:s1_results} (a), our generated contact maps and correspondence maps align well with the object motions, reflecting both physical plausibility and the consistency between object and hand movements. For instance, during the task of opening a book, the corners of the book display higher contact probabilities, indicating that these areas are more likely to be touched.
Additionally, the colors in the embedding maps align with the structure of the hand's embedding (reference Fig.~\ref{fig:embedding}), indicating the geometric and physical plausibility of our results.
For the grasping actions from the GRAB dataset in Fig.~\ref{fig:s1_results} (b), our generated contact maps and embedding maps maintain temporal stability as the object moves, indicating that the hand position remains relatively stable, which is consistent with the nature of a grasping posture.
For the full temporal sequence results, please refer to the \textbf{Sup. Mat.}

\begin{figure*}
    \centering
    \includegraphics[width=1.0\linewidth]{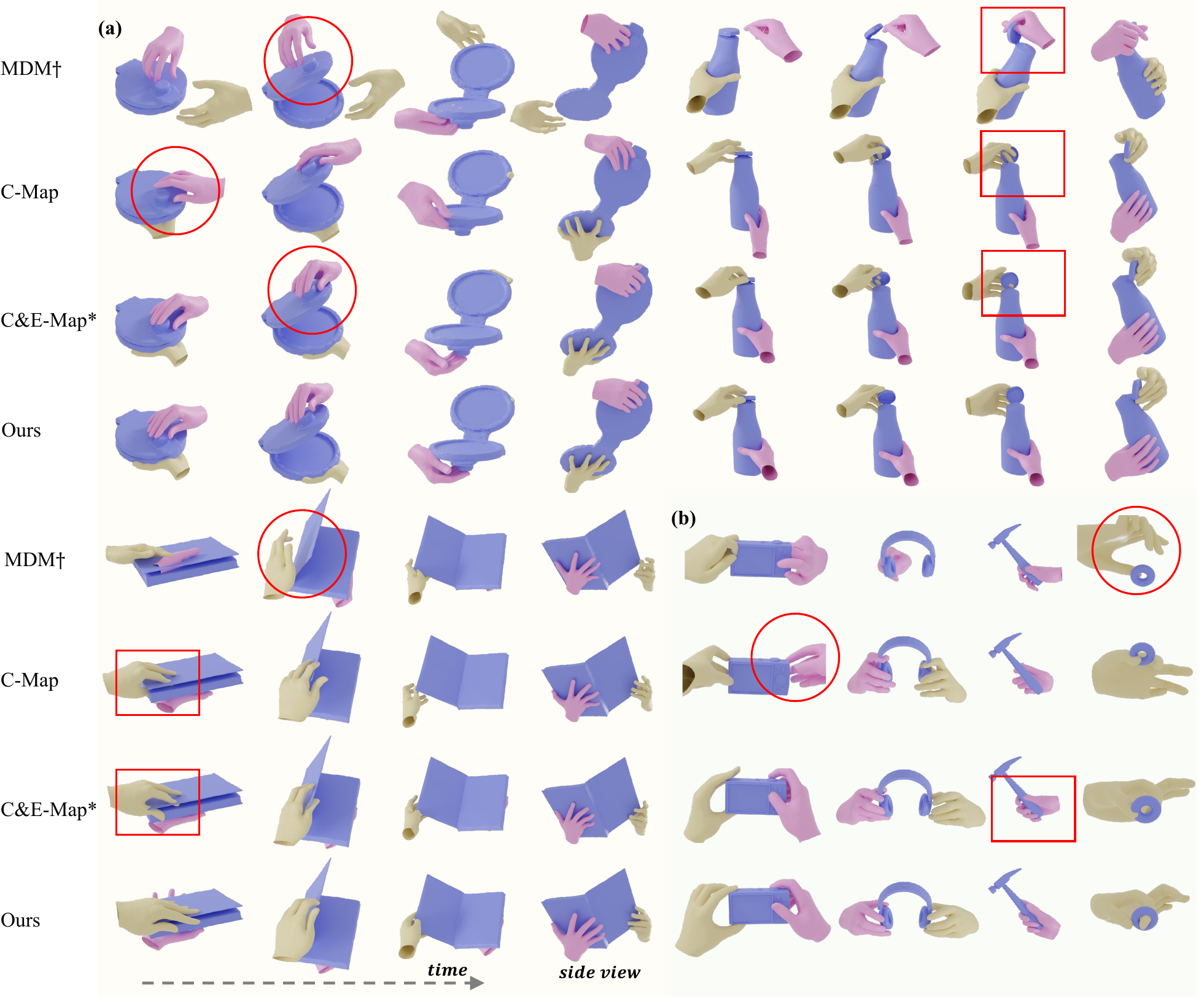}
    \caption{Qualitative Results. We compare our method with baselines (MDM$\dag$, w/ Contact Map, and w/ Contact and Correspondence Embedding Map*) on the ARCTIC (a) and GRAB (b) datasets. The ARCTIC~\cite{arctic} datasets involves manipulations with significant temporal variations, so we visualize several key frames from each sequence. In contrast, the GRAB~\cite{grab} dataset features stable grasps over time, so we visualize only a single frame for brevity.
    For each case, we highlight one representative artifact: circles indicate no contact between hand and object, and squares represent penetration.
    }
    \label{fig:results}
\end{figure*}

\subsection{Grab and Manipulation Synthesis}
\subsubsection{Baselines}
To our knowledge, no previous work has tackled the synthesis of hand manipulations conditioned solely on 3D object trajectories. Existing works either focus on static grasp pose~\cite{contactgen, graspTTA, graspingfield} or dynamic hand grasping conditioned on priors such as body or wrist poses~\cite{manipnet, imos, grab, grip}, hand pose references~\cite{artigrasp, dgrasp, dexmv}, text guidance~\cite{text2hoi}, etc. Considering the prominence of MDM~\cite{mdm} in motion synthesis, we adapt MDM to construct several baselines by replacing its input conditions. The following baselines are established for comparison:
(1) \textbf{MDM$\dag$}. A single-stage framework where the original text conditions are replaced with object trajectories.
(2) \textbf{w/ Contact Map}, C-Map for short. A two stage framework that first generates a contact probability map to guide the subsequent hand pose generation. This baseline demonstrates the effectiveness of a two-stage approach and the benefit of incorporating contact information.
(3) \textbf{w/ Contact and Correspondence Embedding Map}, C\&E-Map for short. This baseline extend C-Map by generating both the contact probability map and a correspondence map in the first stage, allowing us to validate the effectiveness of the proposed continuous correspondence embedding.
(4) \textbf{w/ Contact and Correspondence Embedding Map*}, C\&E-Map* for short. The correspondence map provides the necessary information for calculating $\mathcal{L_\text{att}}$ and $\mathcal{L_\text{consist}}$ losses. Therefore, we further extend C\&E-Map by adding these losses to supervise the correspondence.

\begin{table*}[!t]
    \centering
    \small
    \vspace{2pt}
    \setlength\tabcolsep{.37em}
    \renewcommand{\arraystretch}{1.35}
    \begin{tabular}{l|cccc|cccc}
    \hline
        Dataset & \multicolumn{4}{c|}{ARCTIC} & \multicolumn{4}{c}{GRAB} \\\hline
        Method & Pen $\downarrow$ & V-Contact Ratio $\uparrow$ & V-MPVPE $\downarrow$ & P-Score $\uparrow$ & Pen $\downarrow$ & V-Contact Ratio $\uparrow$ & V-MPVPE $\downarrow$ & P-Score $\uparrow$  \\\hline
        MDM$\dag$~\cite{mdm} & 5.74 & 0.826 & -  & 2.12 & 6.19 & 0.885 & - & 2.59  \\
        C-Map & 6.54 & 0.937 & 3.91 & 2.71 &  5.31 & 0.918 & 3.76 & 3.07  \\
        C\&E-Map & 5.84 & 0.948 & 3.39 & 3.19 & 4.53 & 0.932 & 3.02 & 3.38  \\
        C\&E-Map* & 5.38  & 0.952 & 3.27 & 3.42 & 4.07 & 0.951 & 2.65 & 3.60  \\
        \hline
        Ours (ManiDext) & \textbf{4.65}  & \textbf{0.965}  & \textbf{2.98} & \textbf{3.97} & \textbf{3.70} & \textbf{0.972} & \textbf{2.32} & \textbf{4.02} \\
        \hline
    \end{tabular}
    \caption{Quantitative comparisons between our method and baselines on the ARCTIC~\cite{arctic} and GRAB~\cite{grab} datasets.}
    \label{tab:ablation}
\end{table*}

\subsubsection{Qualitative Results}
Fig.~\ref{fig:results} presents qualitative comparisons between our method and the baseline using the same object trajectories as input. For a fair comparison, all results utilize the same contact probability map and correspondence map generated in the first stage except for the single stage MDM$\dag$.
Consequently,
MDM$\dag$ \ shows frequent penetrations and lack of proper contact with the object. These issues are progressively reduced by introducing contact map, and further mitigated with the help of correspondence map.
Finally, after introducing the residual-guided diffusion module, our full pipeline demonstrates the best results. Specifically, on both the manipulations from the ARCTIC dataset and the grasps from GRAB dataset, our method exhibits more physically plausible interactions with better hand-object motion consistency, improved hand-object contact and fewer penetrations. 
For example, even with challenging objects like the thin surface of a book in ARCTIC or the small torus in GRAB, our method generates realistic results with accurate contact and minimal penetration.
For the full temporal sequence results, please refer to \textbf{Sup. Mat.}

\subsubsection{Quantitative Results}
We split each long motion sequence in the test set  into overlapping 120 frame segments with a stride of 60 frames, ensuring comprehensive coverage of the data. 
For a fair comparison, all results are evaluated using the same contact information generated in the first stage.
As shown in Tab.~\ref{tab:ablation}, our method outperforms other baselines across all objective metrics, including penetration, V-contact ratio, and V-MPVPE, while also achieving the highest participant preference scores in subjective perceptual evaluations on both the ARCTIC and GRAB datasets.
Although MDM$\dag$ shows relatively low penetration, its low contact ratio indicates that the hand does not sufficiently contact the object, leaving a noticeable gap.
By incorporating a two-stage framework and progressively adding contact maps, embedding maps, correspondence loss, and residual guidance module in our method, the results show progressive and significant improvements. 
The results are consistent with our expectations that integrating contact information leads to better performance. 
These experiments also serve as an ablation study, validating the effectiveness of the embedding map and residual guidance module.

\subsubsection{Ablation Study}
\paragraph{Residual Guided Diffusion}
To validate the effectiveness of our proposed residual-guided diffusion module, we compared the results of our method with the baseline \textbf{w/ Contact and Correspondence Embedding Map*}. The difference between these two settings is the introduction of the residual condition. As shown in the qualitative results in Tab.~\ref{tab:ablation} and the quantitative results in Fig.~\ref{fig:results}, the addition of the residual condition improves hand-object contact, reduces penetrations, and enhances performance across all metrics.

\begin{figure}
    \centering
    \includegraphics[width=1.0\linewidth]{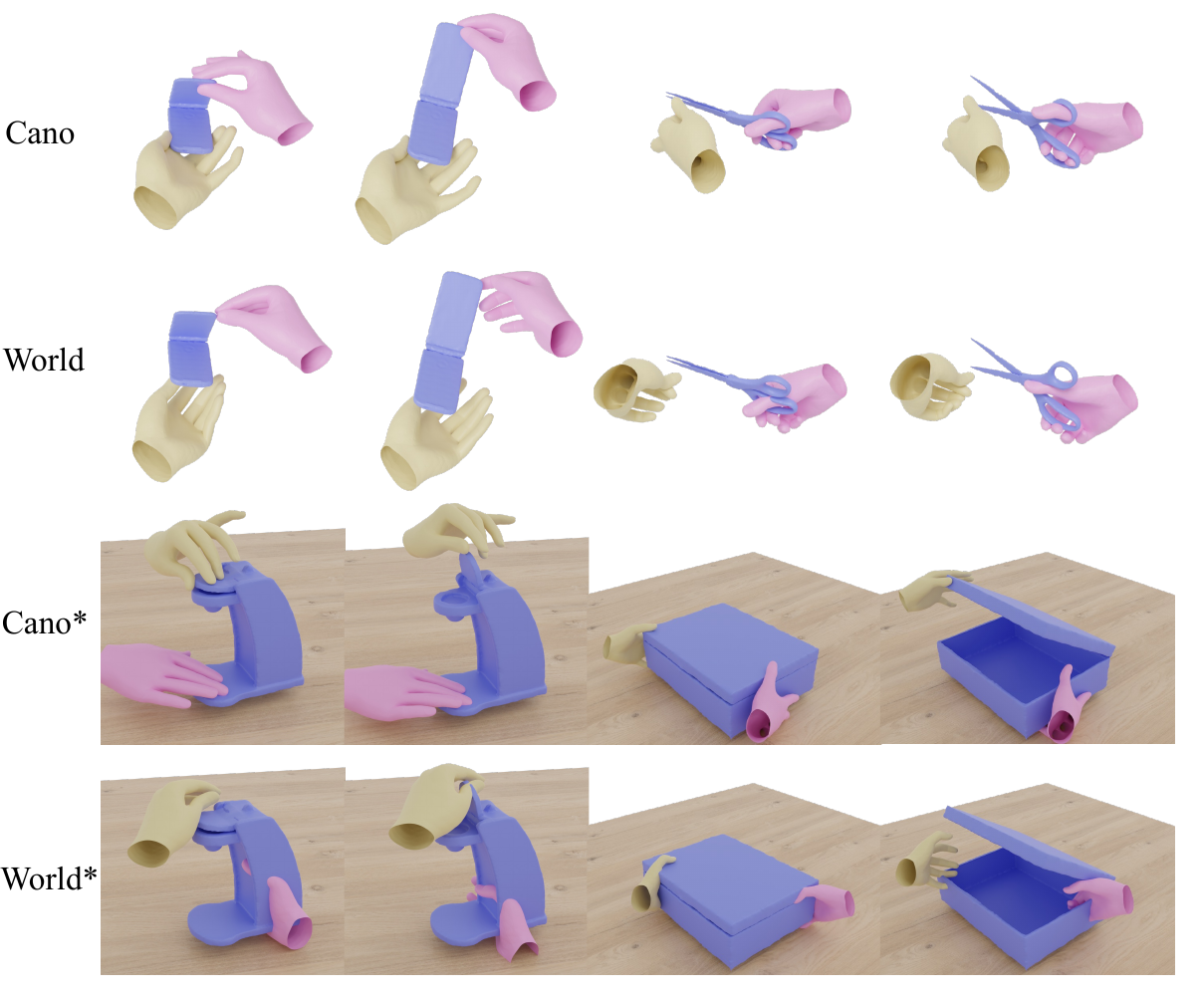}
    \caption{Ablation study on differential object-centric motion modeling. `Cano' represents our model while `World' represents model trained in the world coordinate system. 
    `*' means applying random rotation and translation augmentations to the input object trajectories. 
    }
    \label{fig:ab_cano}
\end{figure}

\paragraph{Differential Object-Centric Motion Modeling}
To evaluate the effectiveness of differential object-centric modeling (Sec.~\ref{sec:object-centric-model}), we trained a model using motion in global world coordinate for comparison. As shown in Tab.~\ref{tab:ablation_study}, our approach in the object's canonical coordinate system outperforms the method in the world coordinate across all metrics. 
Qualitative results are presented in Fig.~\ref{fig:ab_cano}.
The top shows the results on the raw test set, while bottom demonstrates that when random rotations around the z-axis and translations are applied, the model trained in world coordinates fails to generate reasonable poses. This suggests that the network overfits to specific positions rather than learning the fundamental hand-object interaction relationships.

\begin{figure}
    \centering
    \includegraphics[width=1.0\linewidth]{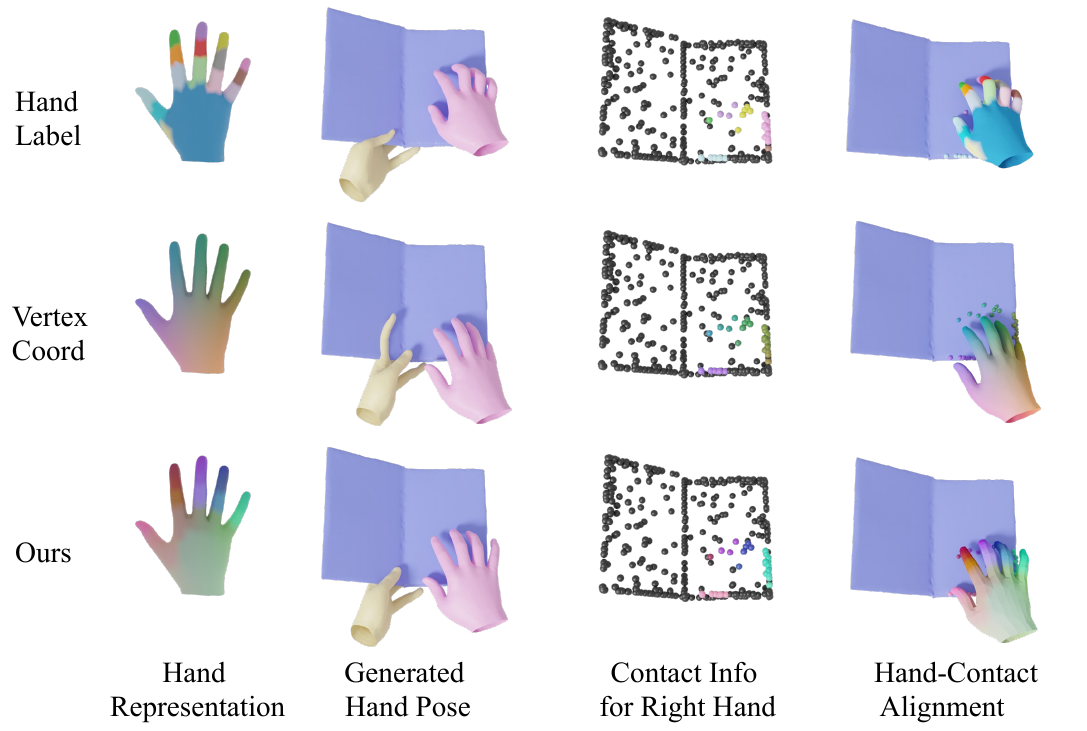}
    \caption{Ablation study of contact representation. Comparison between our continuous correspondence embedding and alternatives using hand part labels or canonical space coordinates. From left to right: hand representation, bimanual results, contact map, and right hand results.}
    \label{fig:ab_emb}
\end{figure}

\paragraph{Continuous Correspondence Embedding} 
To validate the benefit of our proposed continuous correspondence embedding, we compare it with two typical alternative representations of contact information: (1) \textbf{Discrete hand parts label (Part Label)}. We follow ContactGen's~\cite{contactgen} setting which partitions hand model into 16 parts, with each part label associated with a fixed embedding for network training. (2) \textbf{Canonical Vertex Coordinates (Vertex Coord)}. 
Directly using vertex coordinates in the hand's canonical space as an alternative maintains continuity in both the spatial and embedding spaces, while also providing precise vertex-to-vertex correspondence.
Quantitative results in Tab.~\ref{tab:ablation_study} show that our representation outperforms the other two across all metrics. In Fig.~\ref{fig:ab_emb}, we first visualize the hand representations in pseudo color, demonstrating that our method provides clear vertex identity and better distinction among vertices by reflecting geodesic distances rather than Euclidean ones. We further compare the generated hand poses using ground truth contact information for fair comparison, our continuous correspondence embedding not only improves contact accuracy and reduces penetration but also generates poses that are more consistent with the contact information. 
This indicates that our continuous and discriminative representation is more effective for network learning, providing better guidance for pose generation.

\begin{table}[]
    \centering
    \small
    \vspace{2pt}
    \setlength\tabcolsep{.37em}
    \renewcommand{\arraystretch}{1.35}
    \begin{tabular}{l|ccc}
        \hline
        Dataset & \multicolumn{3}{c}{ARCTIC} \\ \hline
        Method & Pen $\downarrow$ & V-Contact Ratio $\uparrow$ & V-MPVPE $\downarrow$ \\\hline
         World Coord & 6.12 & 0.905 & 4.21\\ \hline
         Part Label &  6.07 & 0.941 & 3.60  \\
         Vertex Coord & 6.18 & 0.945 & 3.51 \\ 
         \hline
         ours & \textbf{4.65}  & \textbf{0.965}  & \textbf{2.98} \\
         \hline
    \end{tabular}
    \caption{Ablation study results. We compare different configurations of our method, including training in the world coordinate system and using various contact information representations.}
    \label{tab:ablation_study}
\end{table}

\begin{figure*}
     \centering
    \includegraphics[width=1.0\linewidth]{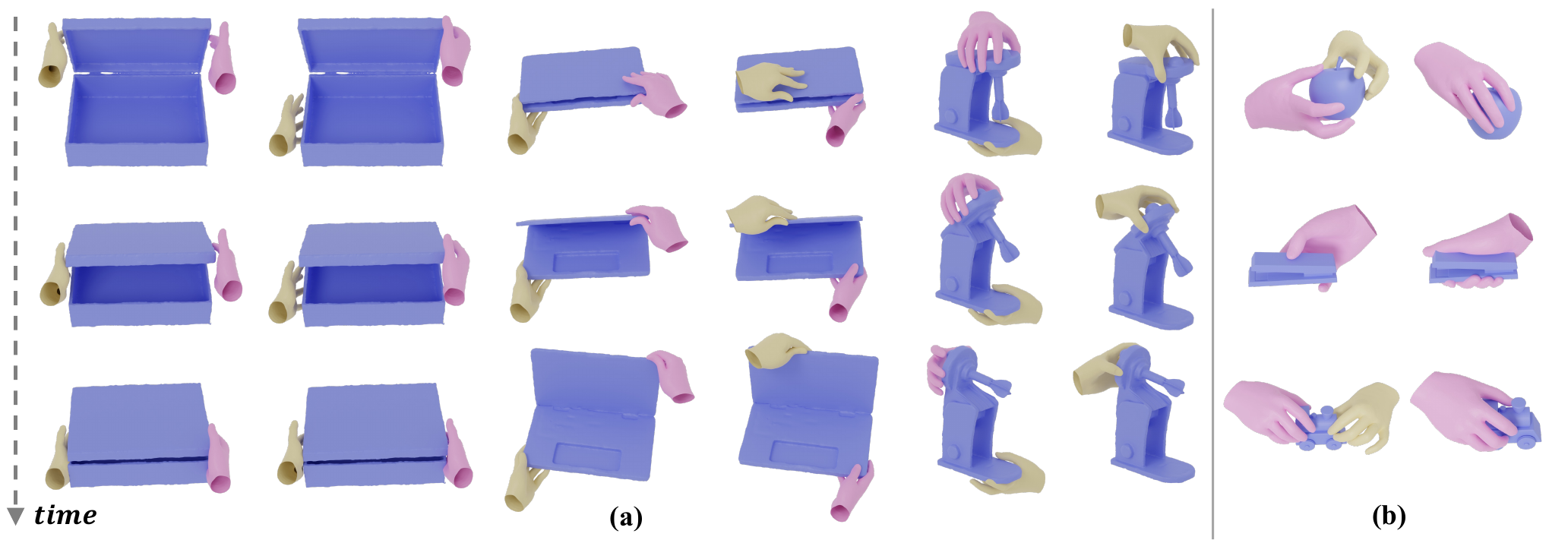}
    \caption{Illustration of the diversity generated by our method. Given the same object trajectory, our method can generate diverse manipulation or grasp poses. (a) shows results from the ARCTIC~\cite{arctic} test set, where key-frames are displayed to highlight the dynamic manipulations. (b) presents results from the GRAB~\cite{grab} test set, where a single frame is shown due to the temporal stability of the grab poses. For both datasets, three cases are shown.} 
    \label{fig:diversity}
\end{figure*}

\subsubsection{Diversity}
In a generative model, diversity is crucial for producing a range of plausible outputs. It's essential for motion generation networks not only to generate realistic and physically plausible motions, but also to capture the variability in the training data. Generating diverse hand poses for the same object motion allows the model to better simulate different manipulation strategies. As shown in  Fig.~\ref{fig:diversity}, given the same object trajectory, our network generates diverse manipulations and grasp poses. These generated hand poses are not only physically plausible but also consistent with the object’s movement. For example, in the ARCTIC dataset, our network generates diverse strategies for behaviors such as opening a box, opening a laptop or operating a mixer, using either single or bimanual hands. Note that, hands that are far away from contacting objects are not shown in Fig.~\ref{fig:diversity}. Similarly, in GRAB dataset, the versatility is shown by grasping objects like an apple, a stapler, or a toy car with either single or bimanual hands. For the same object motion, both manipulation and grasp strategies in Fig.~\ref{fig:diversity} are distinct yet equally plausible. For more results, please refer to \textbf{Sup. Mat.}

\begin{table}[]
    \centering
    \small
    \vspace{2pt}
    \setlength\tabcolsep{.37em}
    \renewcommand{\arraystretch}{1.35}
    \begin{tabular}{ccccccc}
    \hline
    &\multicolumn{3}{c}{Pliers} & \multicolumn{3}{c}{Scissors}   \\ 
    & Pen $(\%)\downarrow$ & Mov $\uparrow$ & Art $\uparrow$ & Pen $(\%)\downarrow$ & Mov $\uparrow$ & Art $\uparrow$ \\ 
    \hline
     GT & 0.000 & 1.000 & 1.000 & 0.046 & 1.000 & 0.970   \\
    \hline
     ManipNet & 0.548 & 0.984 & 0.892 & 0.391 & 0.917 & 0.417  \\
     GraspTTA & 0.555 & 0.779 & 0.420 & 0.454 & 0.993 & 0.849  \\
     CAMS & 0.004 & 1.000 & 1.000 & 0.080 & 0.9999 & 0.989   \\
     \hline
     CAMS- & 0.563 & 0.916 & 0.393 & 0.590 & 0.997 & 0.850   \\
     Ours & 0.354 & 0.939 & 0.976 & 0.190 & 0.969 & 0.964  \\
     \hline
\end{tabular}
\caption{Quantitative results on category-level functional Hand-Object manipulation systhesis task, compared to GraspTTA~\cite{graspTTA}, ManipNet~\cite{manipnet}, CAMS~\cite{cams} and CAMS without optimization (denoted as CAMS-).} 
\label{tab:cams}
\end{table}

\subsection{Category-level Functional Hand-Object Manipulation Synthesis}
Given that our method is the first to generate hand poses conditioned solely on object trajectories, we further evaluate its performance on existing tasks to provide a comprehensive comparison. 
The Category-level Functional Hand-Object Manipulation Synthesis, a novel task introduced by CAMS~\cite{cams}, is defined as generating hand poses for a novel object from a known category, conditioned on a goal sequence that segments the motion into different stages, along with an initial hand pose. 
However, unlike CAMS and other state-of-the-art methods that rely on multiple conditions, we use only object's trajectory as the input condition for comparison. While this poses a greater challenge, it also reduces restrictions during application, making our method more versatile than others.

We follow CAMS, utilizing their released pre-processed HOI4D~\cite{hoi4d} dataset, specifically focusing on the scissors and pliers categories. We employ the same train/test split and consistency metrics, including (1) Contact-Movement Consistency: assessing the alignment between object movement and hand-object contact forces, (2) Articulation Consistency: evaluating whether hand poses manipulate articulated objects in a human-like manner based on torque, and (3) Penetration Rate: the proportion of hand vertices that penetrate the object.  

Quantitative results are presented in Tab.~\ref{tab:cams}, where we compare methods that incorporate explicit optimization (such as ManipNet~\cite{manipnet}, GraspTTA~\cite{graspTTA}, and CAMS~\cite{cams}) against methods without explicit optimization (such as CAMS- and ours). 
As a result, our method outperforms most state-of-the-art methods, 
and show comparable performance to CAMS.
While CAMS, with its explicit optimization approach, can achieve strong numerical performance, its reliance on dense correspondences and the use of separate models for each action-object pair often results in the generation of relatively formulaic and fixed motions, limiting its ability to extend to more dynamic manipulations.
Additionally, the regularization terms used for optimization are not robust or stable across different objects or motions, requiring adjustments for optimal performance, which limits its applicability across various scenarios.
In contrast, our method does not rely on explicit optimization techniques and requires fewer conditions. 
Furthermore, our approach is significantly faster than CAMS, achieving 17 FPS compared to CAMS' 1.6 FPS on the same RTX 3090 device. This highlights our method's efficiency and its potential for broader applications.

\begin{figure}
    \centering
    \includegraphics[width=1.0\linewidth]{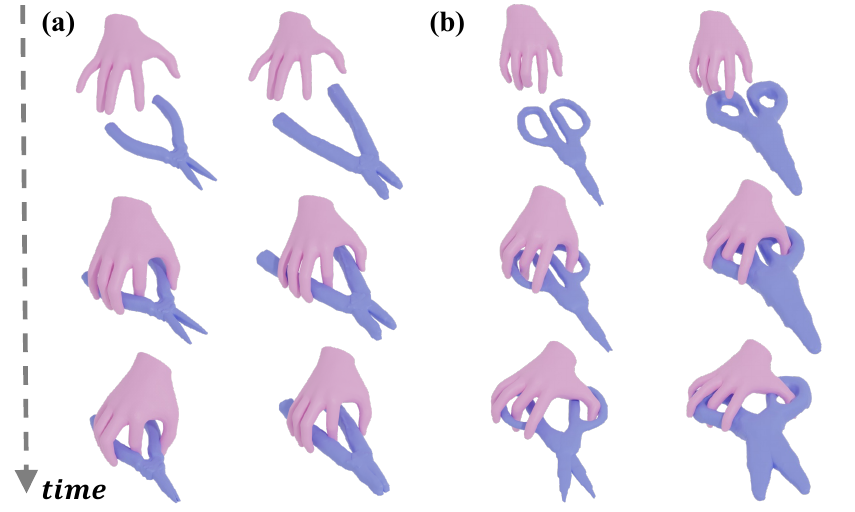}
    \caption{Qualitative results of our method on the HOI4D~\cite{hoi4d} test set for the category-level functional hand-object manipulation synthesis task, showcasing (a) pliers and (b) scissors.}
    \label{fig:cams}
\end{figure}

Qualitative results are shown in Fig.~\ref{fig:cams}, illustrating our method's performance on the test sets for pliers and scissors. We present three key frames for each sequence, representing the approach, grab, and manipulation phases respectively. Despite variations in object shapes within the same category, our method still produces physically plausible interactions consistently, with strong hand-object contact aligned with object motion and minimal penetration.
For each category, the training set includes 7 instances, while the test set comprises of 3 unseen instances. 
These results prove that our method generalizes well to novel instances within the same object category. 

\section{Discussion}
\paragraph{Conclusion}
We propose ManiDext, a hierarchical diffusion framework designed to generate dexterous hand manipulation poses given 3D object trajectories, 
pioneering the bimanual hand generation conditioned solely on object motions. 
To this end, a continuous correspondence embeddings representation is proposed to depict dense contact correspondences between objects and hands, equipped with a residual-guided diffusion module that integrates the iterative refinement process into the diffusion process. These two key algorithmic innovations enable our method to achieve state-of-the-art performance on several object manipulation and grasping datasets, improving the physical fidelity and motion realism of generated hand poses. 
Moreover, these modules can be easily integrated into existing frameworks, whether for single hand or bimanual hand tasks, and whether rigid objects or articulated objects, showcasing significant potential applications in digital humans and embodied AI.

\paragraph{Limitation and Future Works}
Although ManiDext generates physically plausible poses, it ignores possible environment information and may produce conflicting motions, such as attempting to lift an object from below when it is placed on a table. To address this, our future work will integrate upper body constraints and enhance the model's awareness of the scene and object placement. 
While we've shown some generalization ability within the same object categories, the significant variation in motion across different articulated objects makes it challenging to generalize to novel objects from a novel category. A promising solution is to expand the dataset by collecting more real-world data or generating high-quality synthetic data, thus covering a broader range of articulated categories.
Moreover, the principle of our method could be extended to other interaction generation tasks, such as human-object interactions, human-scene interactions, or even human-animal interactions. We hope that our method could inspire more impressive results in the community of motion synthesis. 


\ifCLASSOPTIONcompsoc
  \section*{Acknowledgments}
\else
  \section*{Acknowledgment}
\fi
The work is supported by the National Natural Science Foundation of China (NSFC) under Grant Number 62125107 and 62377004.

\ifCLASSOPTIONcaptionsoff
  \newpage
\fi

\bibliographystyle{unsrt}
\bibliographystyle{IEEEtran}
\bibliography{IEEEabrv,reference}

\begin{thebibliography}{10}

\bibitem{efficientvr}
Markus H{\"o}ll, Markus Oberweger, Clemens Arth, and Vincent Lepetit.
\newblock Efficient physics-based implementation for realistic hand-object interaction in virtual reality.
\newblock In {\em 2018 IEEE conference on virtual reality and 3D user interfaces (VR)}, pages 175--182. IEEE, 2018.

\bibitem{wu2020hand}
Min-Yu Wu, Pai-Wen Ting, Ya-Hui Tang, En-Te Chou, and Li-Chen Fu.
\newblock Hand pose estimation in object-interaction based on deep learning for virtual reality applications.
\newblock {\em Journal of Visual Communication and Image Representation}, 70:102802, 2020.

\bibitem{dexhand}
Haiyan Jiang, Dongdong Weng, Zhen Song, Xiaonuo Dongye, and Zhenliang Zhang.
\newblock Dexhand: dexterous hand manipulation motion synthesis for virtual reality.
\newblock {\em Virtual Reality}, 2023.

\bibitem{imitationrobot}
Anand Thobbi and Weihua Sheng.
\newblock Imitation learning of hand gestures and its evaluation for humanoid robots.
\newblock In {\em The 2010 IEEE International Conference on Information and Automation}, pages 60--65. IEEE, 2010.

\bibitem{deeprobotic}
Tianhao Zhang, Zoe McCarthy, Owen Jow, Dennis Lee, Xi~Chen, Ken Goldberg, and Pieter Abbeel.
\newblock Deep imitation learning for complex manipulation tasks from virtual reality teleoperation.
\newblock In {\em 2018 IEEE international conference on robotics and automation (ICRA)}, pages 5628--5635. IEEE, 2018.

\bibitem{ueda2003handhci}
Etsuko Ueda, Yoshio Matsumoto, Masakazu Imai, and Tsukasa Ogasawara.
\newblock A hand-pose estimation for vision-based human interfaces.
\newblock {\em IEEE Transactions on Industrial Electronics}, 50(4):676--684, 2003.

\bibitem{see2touch}
Irmak Guzey, Yinlong Dai, Ben Evans, Soumith Chintala, and Lerrel Pinto.
\newblock See to touch: Learning tactile dexterity through visual incentives, 2024.

\bibitem{bi-dexhands}
Yuanpei Chen, Yiran Geng, Fangwei Zhong, Jiaming Ji, Jiechuang Jiang, Zongqing Lu, Hao Dong, and Yaodong Yang.
\newblock Bi-dexhands: Towards human-level bimanual dexterous manipulation.
\newblock {\em IEEE Transactions on Pattern Analysis and Machine Intelligence}, 46(5):2804--2818, 2024.

\bibitem{unidexgrasp++}
Weikang Wan, Haoran Geng, Yun Liu, Zikang Shan, Yaodong Yang, Li~Yi, and He~Wang.
\newblock Unidexgrasp++: Improving dexterous grasping policy learning via geometry-aware curriculum and iterative generalist-specialist learning.
\newblock In {\em Proceedings of the IEEE/CVF International Conference on Computer Vision}, pages 3891--3902, 2023.

\bibitem{HInDex}
Yanjie Ze, Yuyao Liu, Ruizhe Shi, Jiaxin Qin, Zhecheng Yuan, Jiashun Wang, and Huazhe Xu.
\newblock H-index: Visual reinforcement learning with hand-informed representations for dexterous manipulation.
\newblock {\em Advances in Neural Information Processing Systems}, 36, 2024.

\bibitem{arctic}
Zicong Fan, Omid Taheri, Dimitrios Tzionas, Muhammed Kocabas, Manuel Kaufmann, Michael~J Black, and Otmar Hilliges.
\newblock Arctic: A dataset for dexterous bimanual hand-object manipulation.
\newblock In {\em Proceedings of the IEEE/CVF Conference on Computer Vision and Pattern Recognition}, pages 12943--12954, 2023.

\bibitem{hoi4d}
Yunze Liu, Yun Liu, Che Jiang, Kangbo Lyu, Weikang Wan, Hao Shen, Boqiang Liang, Zhoujie Fu, He~Wang, and Li~Yi.
\newblock Hoi4d: A 4d egocentric dataset for category-level human-object interaction.
\newblock In {\em Proceedings of the IEEE/CVF Conference on Computer Vision and Pattern Recognition}, pages 21013--21022, June 2022.

\bibitem{obman}
Yana Hasson, Gul Varol, Dimitrios Tzionas, Igor Kalevatykh, Michael~J Black, Ivan Laptev, and Cordelia Schmid.
\newblock Learning joint reconstruction of hands and manipulated objects.
\newblock In {\em Proceedings of the IEEE/CVF conference on computer vision and pattern recognition}, pages 11807--11816, 2019.

\bibitem{fhb}
Guillermo Garcia-Hernando, Shanxin Yuan, Seungryul Baek, and Tae-Kyun Kim.
\newblock First-person hand action benchmark with rgb-d videos and 3d hand pose annotations.
\newblock In {\em Proceedings of the IEEE conference on computer vision and pattern recognition}, pages 409--419, 2018.

\bibitem{oakink2}
Xinyu Zhan, Lixin Yang, Yifei Zhao, Kangrui Mao, Hanlin Xu, Zenan Lin, Kailin Li, and Cewu Lu.
\newblock Oakink2: A dataset of bimanual hands-object manipulation in complex task completion.
\newblock In {\em Proceedings of the IEEE/CVF Conference on Computer Vision and Pattern Recognition}, pages 445--456, 2024.

\bibitem{ho3d}
Shreyas Hampali, Mahdi Rad, Markus Oberweger, and Vincent Lepetit.
\newblock Honnotate: A method for 3d annotation of hand and object poses.
\newblock In {\em Proceedings of the IEEE/CVF conference on computer vision and pattern recognition}, pages 3196--3206, 2020.

\bibitem{OakInk}
Lixin Yang, Kailin Li, Xinyu Zhan, Fei Wu, Anran Xu, Liu Liu, and Cewu Lu.
\newblock Oakink: A large-scale knowledge repository for understanding hand-object interaction.
\newblock In {\em Proceedings of the IEEE/CVF conference on computer vision and pattern recognition}, pages 20953--20962, 2022.

\bibitem{taco}
Yun Liu, Haolin Yang, Xu~Si, Ling Liu, Zipeng Li, Yuxiang Zhang, Yebin Liu, and Li~Yi.
\newblock Taco: Benchmarking generalizable bimanual tool-action-object understanding.
\newblock In {\em Proceedings of the IEEE/CVF Conference on Computer Vision and Pattern Recognition}, pages 21740--21751, 2024.

\bibitem{manus}
Chandradeep Pokhariya, Ishaan~Nikhil Shah, Angela Xing, Zekun Li, Kefan Chen, Avinash Sharma, and Srinath Sridhar.
\newblock Manus: Markerless grasp capture using articulated 3d gaussians.
\newblock In {\em Proceedings of the IEEE/CVF Conference on Computer Vision and Pattern Recognition}, pages 2197--2208, 2024.

\bibitem{chord}
Kailin Li, Lixin Yang, Haoyu Zhen, Zenan Lin, Xinyu Zhan, Licheng Zhong, Jian Xu, Kejian Wu, and Cewu Lu.
\newblock Chord: Category-level hand-held object reconstruction via shape deformation.
\newblock In {\em Proceedings of the IEEE/CVF International Conference on Computer Vision}, pages 9444--9454, 2023.

\bibitem{grab}
Omid Taheri, Nima Ghorbani, Michael~J. Black, and Dimitrios Tzionas.
\newblock {GRAB}: A dataset of whole-body human grasping of objects.
\newblock In {\em European Conference on Computer Vision}, 2020.

\bibitem{contact2grasp}
Haoming Li, Xinzhuo Lin, Yang Zhou, Xiang Li, Yuchi Huo, Jiming Chen, and Qi~Ye.
\newblock Contact2grasp: 3d grasp synthesis via hand-object contact constraint.
\newblock In {\em Proceedings of the Thirty-Second International Joint Conference on Artificial Intelligence, {IJCAI-23}}, pages 1053--1061, 8 2023.

\bibitem{contactopt}
Patrick Grady, Chengcheng Tang, Christopher~D Twigg, Minh Vo, Samarth Brahmbhatt, and Charles~C Kemp.
\newblock Contactopt: Optimizing contact to improve grasps.
\newblock In {\em Proceedings of the IEEE/CVF Conference on Computer Vision and Pattern Recognition}, pages 1471--1481, 2021.

\bibitem{graspTTA}
Hanwen Jiang, Shaowei Liu, Jiashun Wang, and Xiaolong Wang.
\newblock Hand-object contact consistency reasoning for human grasps generation.
\newblock In {\em Proceedings of the IEEE/CVF international conference on computer vision}, pages 11107--11116, 2021.

\bibitem{contactgen}
Shaowei Liu, Yang Zhou, Jimei Yang, Saurabh Gupta, and Shenlong Wang.
\newblock Contactgen: Generative contact modeling for grasp generation.
\newblock In {\em Proceedings of the IEEE/CVF International Conference on Computer Vision}, 2023.

\bibitem{graspingfield}
Korrawe Karunratanakul, Jinlong Yang, Yan Zhang, Michael~J Black, Krikamol Muandet, and Siyu Tang.
\newblock Grasping field: Learning implicit representations for human grasps.
\newblock In {\em 2020 International Conference on 3D Vision (3DV)}, pages 333--344. IEEE, 2020.

\bibitem{halo}
Korrawe Karunratanakul, Adrian Spurr, Zicong Fan, Otmar Hilliges, and Siyu Tang.
\newblock A skeleton-driven neural occupancy representation for articulated hands.
\newblock In {\em 2021 International Conference on 3D Vision (3DV)}, pages 11--21. IEEE, 2021.

\bibitem{contactgrasp}
Samarth Brahmbhatt, Ankur Handa, James Hays, and Dieter Fox.
\newblock {ContactGrasp: Functional Multi-finger Grasp Synthesis from Contact}.
\newblock In {\em 2019 IEEE/RSJ International Conference on Intelligent Robots and Systems (IROS)}, 2019.

\bibitem{manipnet}
H~Zhang, Y~Ye, T~Shiratori, and T~Komura.
\newblock Manipnet: neural manipulation synthesis with a hand-object spatial representation.
\newblock {\em ACM Transactions on Graphics}, 2021.

\bibitem{text2hoi}
Junuk Cha, Jihyeon Kim, Jae~Shin Yoon, and Seungryul Baek.
\newblock Text2hoi: Text-guided 3d motion generation for hand-object interaction.
\newblock In {\em Proceedings IEEE Conference on Computer Vision and Pattern Recognition (CVPR)}, 2024.

\bibitem{cams}
Juntian Zheng, Qingyuan Zheng, Lixing Fang, Yun Liu, and Li~Yi.
\newblock Cams: Canonicalized manipulation spaces for category-level functional hand-object manipulation synthesis.
\newblock In {\em Proceedings of the IEEE/CVF Conference on Computer Vision and Pattern Recognition}, pages 585--594, June 2023.

\bibitem{imos}
Anindita Ghosh, Rishabh Dabral, Vladislav Golyanik, Christian Theobalt, and Philipp Slusallek.
\newblock Imos: Intent-driven full-body motion synthesis for human-object interactions.
\newblock In {\em Eurographics}, 2023.

\bibitem{interhandgen}
Jihyun Lee, Shunsuke Saito, Giljoo Nam, Minhyuk Sung, and Tae-Kyun Kim.
\newblock Interhandgen: Two-hand interaction generation via cascaded reverse diffusion.
\newblock In {\em Proceedings of the IEEE/CVF Conference on Computer Vision and Pattern Recognition}, pages 527--537, 2024.

\bibitem{dgrasp}
Sammy Christen, Muhammed Kocabas, Emre Aksan, Jemin Hwangbo, Jie Song, and Otmar Hilliges.
\newblock D-grasp: Physically plausible dynamic grasp synthesis for hand-object interactions.
\newblock In {\em Proceedings of the IEEE/CVF Conference on Computer Vision and Pattern Recognition}, 2022.

\bibitem{artigrasp}
Hui Zhang, Sammy Christen, Zicong Fan, Luocheng Zheng, Jemin Hwangbo, Jie Song, and Otmar Hilliges.
\newblock {ArtiGrasp}: Physically plausible synthesis of bi-manual dexterous grasping and articulation.
\newblock In {\em International Conference on 3D Vision (3DV)}, 2024.

\bibitem{mano}
Javier Romero, Dimitrios Tzionas, and Michael~J. Black.
\newblock Embodied hands: Modeling and capturing hands and bodies together.
\newblock {\em ACM Transactions on Graphics, (Proc. SIGGRAPH Asia)}, 36(6), 2017.

\bibitem{ddpm}
Jonathan Ho, Ajay Jain, and Pieter Abbeel.
\newblock Denoising diffusion probabilistic models.
\newblock {\em Advances in neural information processing systems}, 33:6840--6851, 2020.

\bibitem{pollard2005physically}
Nancy~S Pollard and Victor~Brian Zordan.
\newblock Physically based grasping control from example.
\newblock In {\em Proceedings of the 2005 ACM SIGGRAPH/Eurographics symposium on Computer animation}, pages 311--318, 2005.

\bibitem{kry2006interaction}
Paul~G Kry and Dinesh~K Pai.
\newblock Interaction capture and synthesis.
\newblock {\em ACM Transactions on Graphics (TOG)}, 25(3):872--880, 2006.

\bibitem{li2007data}
Ying Li, Jiaxin~L Fu, and Nancy~S Pollard.
\newblock Data-driven grasp synthesis using shape matching and task-based pruning.
\newblock {\em IEEE Transactions on visualization and computer graphics}, 13(4):732--747, 2007.

\bibitem{grip}
Omid Taheri, Yi~Zhou, Dimitrios Tzionas, Yang Zhou, Duygu Ceylan, Soren Pirk, and Michael~J. Black.
\newblock {GRIP}: Generating interaction poses using latent consistency and spatial cues.
\newblock In {\em International Conference on 3D Vision ({3DV})}, 2024.

\bibitem{toch}
Keyang Zhou, Bharat~Lal Bhatnagar, Jan~Eric Lenssen, and Gerard Pons-Moll.
\newblock Toch: Spatio-temporal object-to-hand correspondence for motion refinement.
\newblock In {\em European Conference on Computer Vision ({ECCV})}. {Springer}, October 2022.

\bibitem{geneoh}
Xueyi Liu and Li~Yi.
\newblock Geneoh diffusion: Towards generalizable hand-object interaction denoising via denoising diffusion.
\newblock In {\em The Twelfth International Conference on Learning Representations}, 2024.

\bibitem{videodiffusionmodel}
Jonathan Ho, Tim Salimans, Alexey Gritsenko, William Chan, Mohammad Norouzi, and David~J Fleet.
\newblock Video diffusion models.
\newblock {\em Advances in Neural Information Processing Systems}, 35:8633--8646, 2022.

\bibitem{mdm}
Guy Tevet, Sigal Raab, Brian Gordon, Yoni Shafir, Daniel Cohen-or, and Amit~Haim Bermano.
\newblock Human motion diffusion model.
\newblock In {\em The Eleventh International Conference on Learning Representations}, 2023.

\bibitem{mld}
Xin Chen, Biao Jiang, Wen Liu, Zilong Huang, Bin Fu, Tao Chen, and Gang Yu.
\newblock Executing your commands via motion diffusion in latent space.
\newblock In {\em Proceedings of the IEEE/CVF Conference on Computer Vision and Pattern Recognition}, pages 18000--18010, 2023.

\bibitem{textdiffusion}
Xiang Li, John Thickstun, Ishaan Gulrajani, Percy~S Liang, and Tatsunori~B Hashimoto.
\newblock Diffusion-lm improves controllable text generation.
\newblock {\em Advances in Neural Information Processing Systems}, 35:4328--4343, 2022.

\bibitem{audiodiffusion}
Nanxin Chen, Yu~Zhang, Heiga Zen, Ron~J Weiss, Mohammad Norouzi, and William Chan.
\newblock Wavegrad: Estimating gradients for waveform generation.
\newblock {\em arXiv preprint arXiv:2009.00713}, 2020.

\bibitem{interdiff}
Sirui Xu, Zhengyuan Li, Yu-Xiong Wang, and Liang-Yan Gui.
\newblock Interdiff: Generating 3d human-object interactions with physics-informed diffusion.
\newblock In {\em Proceedings of the IEEE/CVF International Conference on Computer Vision}, pages 14928--14940, 2023.

\bibitem{omomo}
Jiaman Li, Jiajun Wu, and C~Karen Liu.
\newblock Object motion guided human motion synthesis.
\newblock {\em ACM Transactions on Graphics (TOG)}, 42(6):1--11, 2023.

\bibitem{cghoi}
Christian Diller and Angela Dai.
\newblock Cg-hoi: Contact-guided 3d human-object interaction generation.
\newblock In {\em Proceedings of the IEEE/CVF Conference on Computer Vision and Pattern Recognition}, pages 19888--19901, 2024.

\bibitem{chois}
Jiaman Li, Alexander Clegg, Roozbeh Mottaghi, Jiajun Wu, Xavier Puig, and C~Karen Liu.
\newblock Controllable human-object interaction synthesis.
\newblock {\em arXiv preprint arXiv:2312.03913}, 2023.

\bibitem{hoidiff}
Xiaogang Peng, Yiming Xie, Zizhao Wu, Varun Jampani, Deqing Sun, and Huaizu Jiang.
\newblock Hoi-diff: Text-driven synthesis of 3d human-object interactions using diffusion models.
\newblock {\em arXiv preprint arXiv:2312.06553}, 2023.

\bibitem{bps}
Sergey Prokudin, Christoph Lassner, and Javier Romero.
\newblock Efficient learning on point clouds with basis point sets.
\newblock In {\em Proceedings of the IEEE/CVF international conference on computer vision}, pages 4332--4341, 2019.

\bibitem{pose6d}
Yi~Zhou, Connelly Barnes, Jingwan Lu, Jimei Yang, and Hao Li.
\newblock On the continuity of rotation representations in neural networks.
\newblock In {\em Proceedings of the IEEE/CVF conference on computer vision and pattern recognition}, pages 5745--5753, 2019.

\bibitem{cse}
Natalia Neverova, David Novotny, Marc Szafraniec, Vasil Khalidov, Patrick Labatut, and Andrea Vedaldi.
\newblock Continuous surface embeddings.
\newblock {\em Advances in Neural Information Processing Systems}, 33:17258--17270, 2020.

\bibitem{transformer}
Ashish Vaswani, Noam Shazeer, Niki Parmar, Jakob Uszkoreit, Llion Jones, Aidan~N Gomez, {\L}ukasz Kaiser, and Illia Polosukhin.
\newblock Attention is all you need.
\newblock {\em Advances in neural information processing systems}, 30, 2017.

\bibitem{dexmv}
Yuzhe Qin, Yueh-Hua Wu, Shaowei Liu, Hanwen Jiang, Ruihan Yang, Yang Fu, and Xiaolong Wang.
\newblock Dexmv: Imitation learning for dexterous manipulation from human videos.
\newblock In {\em European Conference on Computer Vision}, pages 570--587. Springer, 2022.

\end{thebibliography}
\vspace{-12mm}

\end{document}